\documentclass[sigconf]{acmart}
\usepackage[utf8]{inputenc}
\usepackage[super]{nth}
\usepackage{amssymb,amsmath,amsfonts,amsthm,enumitem,bm}
\usepackage{verbatim}
\usepackage{subcaption}
\usepackage{url}
\usepackage{array}

\expandafter\def\expandafter\normalsize\expandafter{%
    \normalsize
    \setlength\abovedisplayskip{0pt}
    \setlength\belowdisplayskip{1pt}
    \setlength\abovedisplayshortskip{-5pt}
    \setlength\belowdisplayshortskip{4pt}
}

\DeclareMathOperator{\topk}{topk}
\DeclareMathOperator{\sign}{sgn}

\newcommand{\HF}{HF}

\newcommand{\oneshot}{one-shot}
\newcommand{\Oneshot}{One-shot}
\newcommand{\oneshotGen}{\oneshot\ classification}
\newcommand{\OneshotGen}{\Oneshot\ classification}
\newcommand{\oneshotInst}{\oneshot\ instance-classification}
\newcommand{\OneshotInst}{\Oneshot\ instance-classification}

\newcommand{\instance}{instance}
\newcommand{\PC}{PC(CA3)}
\newcommand{\PS}{PS(DG)}
\newcommand{\PR}{PR(EC-CA3)}
\newcommand{\VC}{VC(EC)}
\newcommand{\PM}{PM(CA1)}
\newcommand{\FNN}{FastNN}

\begin{document}

\title{AHA! an `Artificial Hippocampal Algorithm' for Episodic Machine Learning}

\author{Gideon Kowadlo}
\affiliation{Incubator 491}
\email{gideon@agi.io}

\author{Abdelrahman Ahmed}
\affiliation{Incubator 491}
\email{abdel@agi.io}

\author{David Rawlinson}
\affiliation{Incubator 491}
\email{dave@agi.io}

\date{March 2020}

\begin{abstract}
The majority of ML research concerns slow, statistical learning of i.i.d. samples from large, labelled datasets. Animals do not learn this way. An enviable characteristic of animal learning is `episodic' learning - the ability to memorise a specific experience as a composition of existing concepts, after just one experience, without provided labels. The new knowledge can then be used to distinguish between similar experiences, to generalise between classes, and to selectively consolidate to long-term memory.
The Hippocampus is known to be vital to these abilities.
AHA is a biologically-plausible computational model of the Hippocampus. Unlike most machine learning models, AHA is trained without external labels and uses only local credit assignment.
We demonstrate AHA in a superset of the Omniglot \oneshotGen\ benchmark. The extended benchmark covers a wider range of known hippocampal functions by testing pattern separation, completion, and recall of original input.
These functions are all performed within a single configuration of the computational model. Despite these constraints, image classification results are comparable to conventional deep convolutional ANNs.

\end{abstract}

\maketitle

\section{Introduction}
\label{sec:introduction}


In recent years, machine learning has been applied to great effect across many problem domains including speech, vision, language and control \citep{Lake2017, Vinyals2016}. The dominant approach is to combine models having a very large number of trainable parameters with careful regularisation and slow, statistical learning. Samples are typically assumed to be i.i.d. (independent and identically distributed), implying an unchanging world.
Popular algorithms require large volumes of data, and with the exception of some very recent image classification meta-learning frameworks e.g. \citep{Berthelot2019, Sohn2020}, an equally large number of labels. Models are usually susceptible to ``catastrophic forgetting'', without an ability to learn continually, also referred to as lifelong learning \citep{Kirkpatrick2017, Sodhani2020}.

In contrast, animals display a much broader range of intelligence \citep{Lake2017} seemingly free of these constraints.
A significant differentiator between animals and machines, is the ability to learn from one experience (\oneshot\ learning), to reason about the specifics of an experience, and to generalise learnings from the one experience.
In addition, new knowledge exploits the compositionality of familiar concepts \citep{Lake2017}, which is likely a key component for continual learning. For example, after seeing a single photo of a giraffe with a birthmark on its ear, a young child can recognise other giraffes and identify this specific one as well.\footnote{Based on an example by \cite{Vinyals2016}.}

\Oneshot\ learning of this nature is of great interest to better understand animal intelligence as well as to advance the field of machine intelligence. 

\subsection{Mammalian Complementary Learning Systems}
\begin{figure}[ht]
  \centering
    \includegraphics[width=0.7\columnwidth]{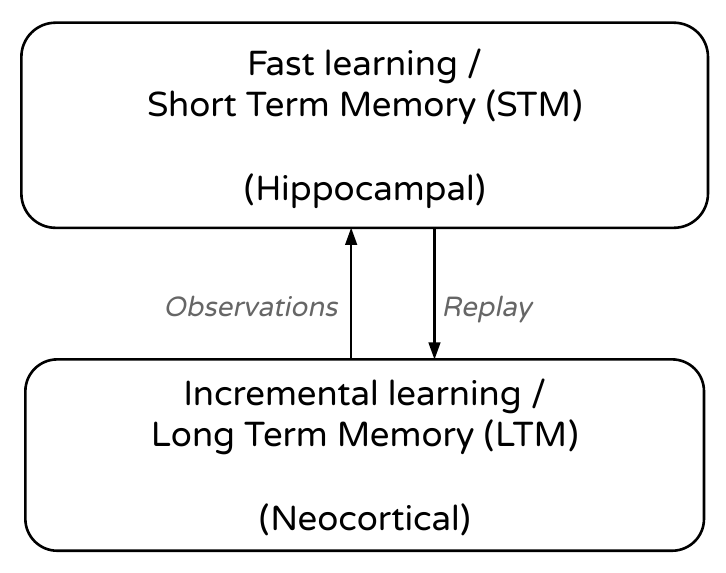}
  \caption{\textbf{High-level depiction of Complementary Learning Systems (CLS)}. The specialisations of two memory systems complement each other. The short-term memory learns and forgets rapidly; salient memories can be replayed to the long-term memory, enabling incremental statistical learning there.}\label{fig:complementary_systems}
\end{figure}

In mammals, a brain region widely recognised to be critical for these aspects of learning and memory is the Hippocampal Formation (\HF) \citep{Kandel1991}.\footnote{The hippocampus itself is contained within the \HF\ along with other structures (there is no universally accepted definition of the exact boundaries). The \HF\ is often described as being contained within a region called the Medial Temporal Lobe (MTL).}.
Complementary Learning Systems (CLS) is a standard framework for understanding the function of the \HF\ \citep{McClelland1995, OReilly2014, Kumaran2016}. CLS consists of two differentially specialised and complementary structures, the neocortex and the \HF. They are depicted in a high level diagram in Figure~\ref{fig:complementary_systems}.
In this framework, the neocortex is most similar to a conventional ML model, incrementally learning regularities across many observations comprising a long-term memory (LTM). It forms overlapping and distributed representations that are effective for inference. In contrast, the \HF\ rapidly learns distinct observations, forming sparser, non-overlapping and therefore non-interfering representations, functioning as a short term memory (STM). Recent memories from the \HF\ are replayed to the neocortex, re-instating the original activations resulting in consolidation as long-term memory (LTM). Patterns are replayed in an interleaved fashion, avoiding catastrophic forgetting. In addition, they can be replayed selectively according to salience. 
There have been numerous implementations of CLS \citep{Norman2003, Ketz2013, Greene2013, Rolls2013, Schapiro2017a} and \cite{Rolls1995} published a similar model with greater neuroanatomical detail.
A more detailed description of the \HF\ is given in Section~\ref{sec:comp_model}.

CLS explains the role of the \HF\ in Episodic Memory \citep{Gluck2003}. In this and related literature (citations above), an episode is defined as a distinct observation that is a ``conjunctive binding of various distributed pieces of information'' \citep{Ketz2013}.
It is more general than the common behavioural definition introduced by \cite{Tulving1985}, an (often autobiographical) memory of places and things, ``memories for the specific contents of individual episodes or events'' \citep{McClelland1995}. 
In the behavioural definition, a temporal aspect with ordering is implied, but is not necessary.
CLS-style learning is thus a potential bridge to endowing machines with Episodic Memory capabilities \citep{Kumaran2016}. 
\footnote{Framed as a conjunction of concepts, \oneshot\ learning of episodes also provides the tantalising potential to learn factual  information, traditionally defined as Semantic Memory \citep{Squire1992}. The memories held temporarily in the \HF\ (STM) and consolidated in the neocortex (LTM) could consist of both Episodic and Semantic Memory, together comprising Declarative Memory \citep{McClelland1995, Rolls2013, Goldberg2009}.}

\subsection{Motivation and Solution}
The motivation of this work is to understand and replicate \HF\ functionality to reproduce the advantages of this form of \oneshot\ learning.
We are therefore focussing on the functional capabilities of Episodic Memory as described by CLS. The \HF\ also performs other functions that are out of scope for this study, such as incremental learning \citep{Gluck2003, Schapiro2017a} and prediction \citep{Lisman2009, Buckner2010, Stachenfeld2017}. It is also known for its role in spatial mapping using Place and Grid cells, but is understood to be more general conceptual mapping rather than being confined to purely spatial reasoning \citep{Moser2008, Kurth-Nelson2016, Mok2019}.

The CLS approach (citations above) is to model neuron dynamics and learning rules that are biologically realistic at the level of the individual neuron.
As a consequence, the behaviour of the region is limited by the accuracy of the neuron model. In addition, it may be more difficult to scale than collectively modelling populations of neurons at a higher level of abstraction. Scaling may be important to work with inputs derived from realistic sensor observations such as images and compare to conventional ML models.
Prior CLS experiments demonstrate an ability to recognise specific synthetically generated vectors requiring pattern separation between similar inputs, often with the presence of distractor patterns or `lures'. 
In most cases, generalisation performance is not reported. \citet{Greene2013} introduced noise (additive and non-additive) and occlusion to test inputs, and to our knowledge, no studies have tested the ability to generalise to different exemplars of the same class.

Within the ML literature, \oneshot\ learning has received recent attention. Following seminal work by \cite{Fei-Fei2003, Fei-Fei2006} the area was re-invigorated by \cite{Lake2015} who introduced a popular test that has become a standard benchmark. It is a \oneshot\ classification task on the Omniglot dataset of handwritten characters (see Section~\ref{sec:experimental_method} for a detailed description).

A typical approach is to train a model on many classes and use learnt concepts to recognise new classes quickly from one or few examples. Often framed as meta-learning, or ``learning to learn'', solutions have been implemented with neural networks that use external labels at training time such as Siamese networks \citep{Koch2015a}, matching networks \citep{Vinyals2016}, and prototypical networks \citep{Snell2017}, 
as well as Bayesian approaches \citep{Lake2015, George2017}. A comprehensive review is given in \citep{Lake2019}.
In contrast to the computational neuroscience models, these studies focus on generalisation without considering additional \HF\ capabilities such as modelling of specific instances and pattern separation, and without a framework for retaining knowledge beyond the classification task itself. They do not consider imperfect samples due to factors such as occlusion or noise.

Moving beyond the current CLS and ML \oneshot\ studies requires developing the ability to reason about specific instances in addition to generalising classes. A practical realisation of these capabilities implies pattern separation and completion in addition to learning the observational invariances of classes. These appear to be contradictory qualities, making it challenging to implement both in a unified architecture. It is also challenging to model the \HF\ at an appropriate level of abstraction - detailed enough to capture the breadth of capabilities whilst abstract enough to scale to convincing experiments with realistic sensor data.

We developed an Artificial Hippocampal Algorithm (AHA) to extend \oneshot\ learning in ML and provide insight on hippocampal function.
AHA is a new computational model that can work with realistic sensor data without labels. AHA combines features from CLS and \cite{Rolls1995, Rolls2013}, modelling the hippocampus at the level of subfields (anatomically and physiologically distinct regions). The micro-architecture of AHA is implemented with scalable ML frameworks, subject to biological plausibility constraints (see Section~\ref{sec:bio_plaus}).
Functionally, AHA provides a fast learning STM that complements an incrementally learning LTM model (such as a conventional artificial neural network).
The complete system is tested on a dataset resembling realistic grounded sensor data including the introduction of typical input corruption simulated with noise and occlusion. The tasks involve the learning of specific instances, generalisation and recall for replay to the LTM. We propose a new benchmark that extends the Omniglot test \citep{Lake2015} for this purpose. The experimental method and benchmark are described in detail in Section~\ref{sec:experimental_method}.
The performance of AHA is compared to two baselines: a) the LTM on its own, and b) an STM implemented with standard ML components. The differing qualities of AHA's internal signals are also reported.

\subsection{Contributions}
Our primary contributions are as follows: 
\begin{enumerate}
  \item We propose a novel hippocampal model called AHA, based on CLS, but with a different model for combining pathways, and modelling neuron populations at a higher level of abstraction.
  \item We propose a \oneshot\ learning benchmark that extends the standard Omniglot test to capabilities not typically seen in the ML literature or replicated by hippocampal models in the computational neuroscience literature.
  \item We evaluate LTM+AHA on the proposed benchmark, and show that it is better suited to \oneshot\ learning without labels than the visual processing of LTM alone and a baseline STM. In addition we show that on the base Omniglot \oneshot\ classification test, it is comparable to state-of-the-art neural network approaches that do not contain domain specific priors.
  \item We also show that AHA is a viable complement to an incremental learning LTM, with the ability to store and replay memories when presented with a cue, whereas the baseline STM architecture was less effective.
\end{enumerate}

\section{Model}
\label{sec:model}

In this section, we first describe the biological plausibility constraints that govern the architecture micro-structure. Next, we describe the standard CLS-style model, explaining the biological function and the salient features that are used as the conceptual foundation for AHA. 
The training/testing framework that we used for the LTM+STM architecture is described, providing context for sub-sections on two STM implementations: AHA itself and \FNN, a baseline model. AHA's theory of operation is also discussed.

\subsection{Biological Plausibility Constraints}
\label{sec:bio_plaus}

We adopt the local and immediate credit assignment constraints from \citep{Rawlinson2019}. Under these constraints it is possible to use stochastic gradient descent, but with only shallow (and local) backpropagation. In addition we do not require any labels for training. We do not claim these criteria are sufficient for an accurate replica of biology. However, we aim to avoid the most implausible features of conventional machine learning approaches. 

Our definition of ``shallow'' is to allow error backpropagation across at most two ANN layers.
The computational capabilities of single-layer networks are very limited, especially in comparison to two-layer networks. Biological neurons perform ``dendrite computation'', involving integration and non-linearities within dendrite subtrees \citep{Guerguiev2017, Tzilivaki2019}. This is computationally equivalent to 2 or 3 typical artificial neural network layers. We assume that shallow backpropagation could approximate dendrite computation within a single biological cell layer, and training signals propagated inside cells.

\subsection{Biological Computational Model}
\label{sec:comp_model}

\begin{figure}[ht]
  \centering
    \includegraphics[width=0.98\columnwidth]{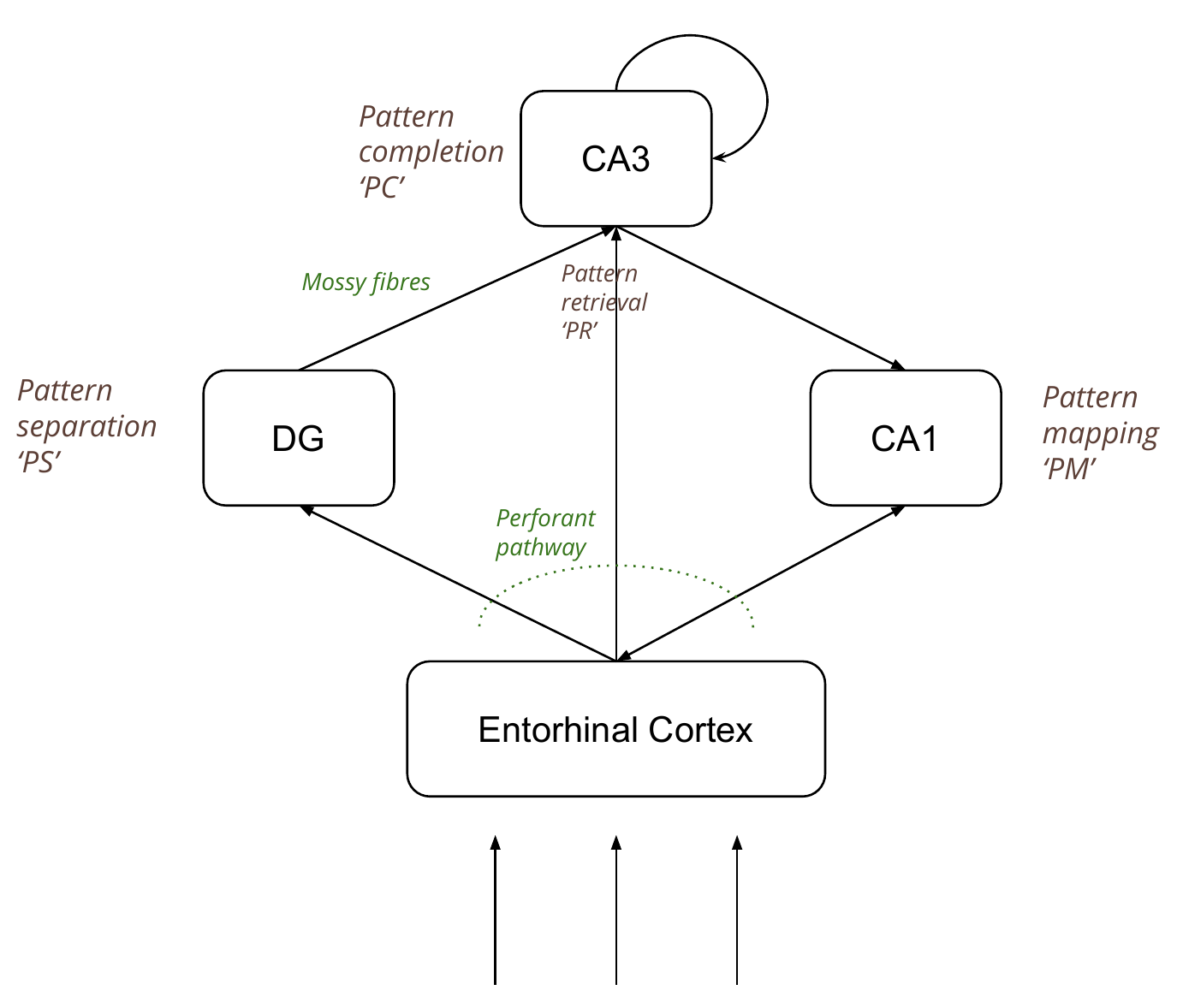}
  \caption{Biological subfields of the \HF.}
  \label{fig:biological_fields}
\end{figure}

The adopted biological model of the \HF\ is based on the CLS-style models of \citep{Rolls1995, Rolls2013, Rolls2018} and CLS itself \citep{McClelland1995, OReilly2014}, illustrated in Figure~\ref{fig:biological_fields}. 
The concept of complementary learning systems was given in the Introduction (Section~\ref{sec:introduction}).
In this section, we describe the subfields (functional units) and pathways of the model, referring to the citations given above. 
We choose details and a level of description relevant to understanding and implementing functionality in a working model. 

The Entorhinal Cortex (EC) is the main gateway between the neocortex and hippocampus.
It does not fit neatly into the overly simplified model shown in Figure~\ref{fig:complementary_systems}, as it is broadly part of the \HF, but thought to learn incrementally \citep{Gluck2003}.
Entorhinal cortex sends input from superficial layers (EC\textsubscript{in}) to the hippocampus, and deep layers (EC\textsubscript{out}) receive output from the hippocampus. 
The hippocampus learns to retrieve and `replay' EC\textsubscript{in} patterns to EC\textsubscript{out}, where they are reinstated (equivalent to `reconstruction' in ML terminology). 
At a high level, the hippocampus comprises an auto-associative memory that can operate effectively with partial cues (pattern completion) and distinguish very similar inputs (pattern separation).

Within the hippocampus, there are several functional units or subfields. The most significant of which, according to CLS and Rolls, are DG, CA3 and CA1.
EC\textsubscript{in} forms a sparse and distributed overlapping pattern that combines input from all over the neocortex and subcortical structures. This pattern becomes sparser and less overlapping through Dentate Gyrus (DG) and CA3 with increasing inhibition and sparse connectivity. 
That provides distinct representations for similar inputs and therefore an ability to separate patterns, important for learning about specific episodes or conjunctions of input concepts, as opposed to generalities. 
The DG-CA3 connections (Mossy fibres) are non-associative and are responsible for encoding engrams in CA3 \citep{Rolls2013}. 
The EC-CA3 connections comprise a pattern association network and are responsible for providing a cue for retrieval (also referred to as recall) from CA3 \citep{Rolls2013}. 
Recurrent connections in CA3 create an auto-associative memory with basins of attraction storing multiple patterns. Any part of a stored pattern can be presented as a cue to recall a crisp and complete version of the closest pattern. 
EC has bilateral connections to CA1 and the CA3-CA1-EC pathway forms a hierarchical pattern associative network. 
During encoding, activated CA1 neurons form associative connections with the active EC neurons.
During retrieval, the sparse CA3 pattern becomes denser and more overlapping through CA1, resulting in the original, complete pattern that was present during encoding being replayed, reinstating activation in EC\textsubscript{out} and in turn neocortex through reciprocal feedback connections.
The replay is used for cognitive processing or for long-term memory consolidation. 
Replay occurs in an interleaved fashion, allowing incremental learning in the neocortex without catastrophic forgetting.

Electrophysiological recordings and lesion studies show that oscillatory activity called theta rhythms, correspond to encoding and retrieval phases \citep{Hasselmo2002, Ketz2013}. During encoding, synaptic transmission from EC is strong and from CA3 weak, although the CA3 and CA1 synapses show high potential for synaptic modification (plasticity). During retrieval, synaptic transmission from CA3 is strong, and from EC is weak but sufficient to provide a cue, facilitating retrieval. For this reason our model has discrete encoding (training) and retrieval (inference) phases, modulated by the respective subfield equivalents.

In the CLS-style studies mentioned above and subsequent implementations \citep{Norman2003, Ketz2013, Greene2013, Schapiro2017a}, DG and EC projections are involved in encoding and retrieval in CA3. In contrast, \cite{Rolls1995, Rolls2013, Rolls2018} presents evidence that encoding in CA3 is dominated by strong afferent activity from DG facilitated by non-associative plasticity. Whereas, retrieval with generalisation takes place through associative learning in EC-CA3. 
We adopt Rolls' model, as DG's highly orthogonal patterns are effectively random and are likely to `confuse' cueing CA3.

\begin{figure}[ht]  
  \centering
    \includegraphics[width=0.75\columnwidth]{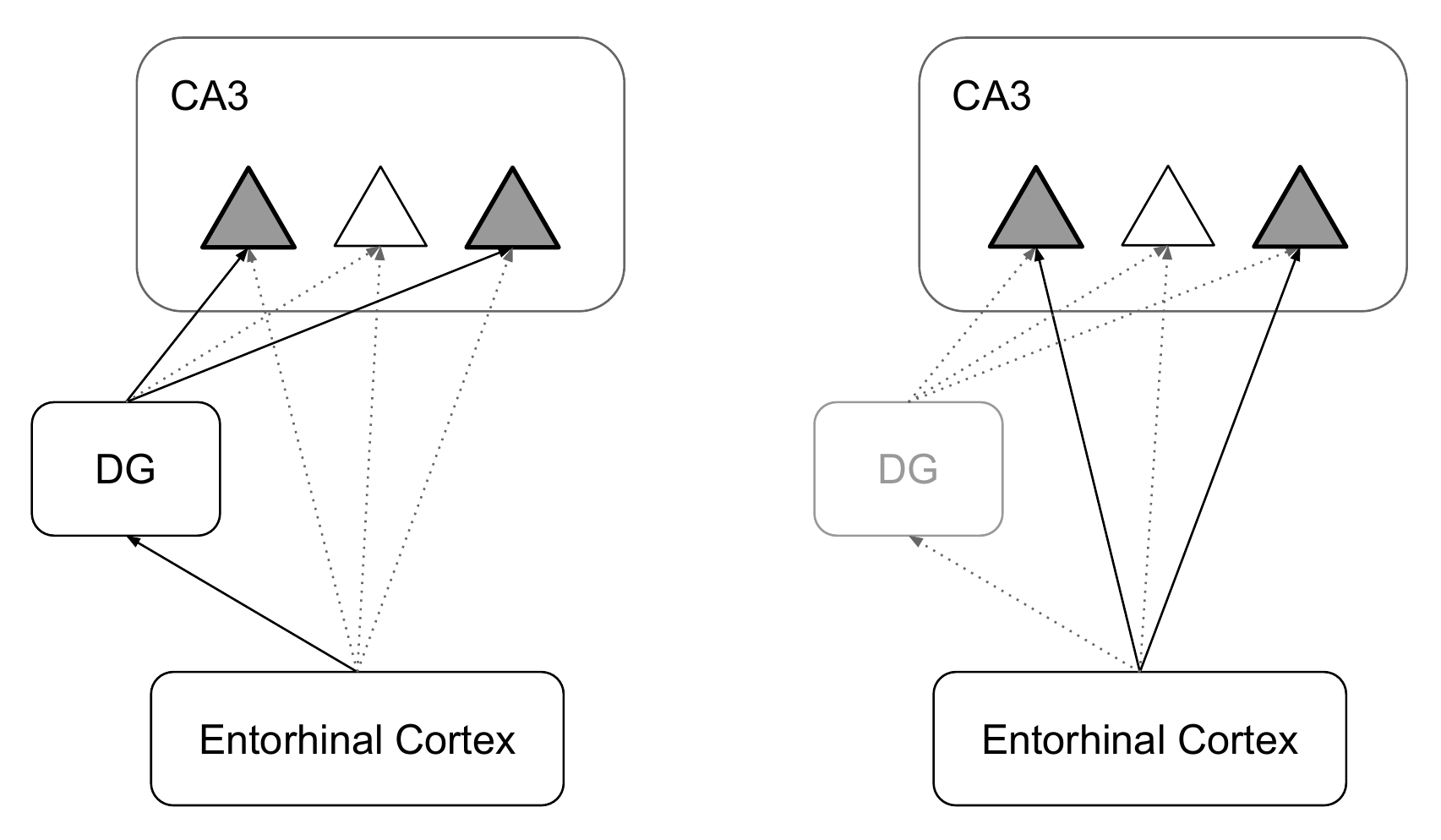}
  \caption{\textbf{Encoding and retrieval in CA3.} Left: An EC pattern flows through DG and results in a sparse set of active neurons in CA3 (active neurons are shown filled with grey, inactive are white). Right: Simultaneously, the same EC pattern forms associative connections between active neurons in EC and those in CA3 which represent the stored pattern from DG. Solid arrows represent axon firing, dotted arrows represent inactive axons.}
  \label{fig:pr_connectivity}
\end{figure}

Roll's model of encoding and retrieval with CA3 \citep{Rolls1995, Rolls2013, Rolls2018} is expanded here and illustrated in Figure~\ref{fig:pr_connectivity}.
During encoding, strong DG transmission results in a pattern of active pyramid neurons in CA3, comprising an engram that becomes an attractor state through synaptic modification in the recurrent neurons.
EC-CA3 concurrently learns to associate EC and CA3 representations.
In effect, the CA3 engram comprises a target for later retrieval by EC-CA3, that can be used to cue recall from CA3. Since EC representations are distributed and overlapping, similar observations are able to retrieve the same appropriate distinct memory, enabling generalisation.

Conversely, since the CA3 engram varies greatly in response to small changes in EC, the EC-DG-CA3 pathway enables separation of similar patterns. However, this also prevents generalisation via this pathway. Together, the two pathways EC-DG-CA3 and EC-CA3 are able to perform the opposing qualities of pattern generalisation and pattern separation.

More recent versions of CLS \citep{Greene2013, Ketz2013, Schapiro2017a} utilise biologically plausible Contrastive Hebbian Learning (CHL) within the LEABRA framework \citep{OReilly1996}. This form of Hebbian learning uses local error gradients, speeding up training and avoiding a pre-training step. The use of local error gradients is consistent with our biological constraints.

`Big-loop' recurrence occurs when a reinstated EC pattern is transmitted back into the \HF\ as the EC input. This process has received little attention in the computational modelling literature. It appears to play a role in learning statistical information in the form of higher order patterns across episodes \citep{Schapiro2017a, Koster2018} and is therefore out of scope for this study.

\subsection{Training and Testing Framework}
\label{sec:train_test_fwk}

LTM and STM components (Figure~\ref{fig:complementary_systems}) are trained and tested at different stages. This section lays out the framework providing context for the model descriptions. In the following sections related to ML implementations, we use the term `Train(ing)' to refer to the process of encoding and `Test(ing)' to refer to retrieval/recall. The stages of testing and training are described below.

\noindent \textbf{Stage 1: Pre-train LTM}: Train LTM on a disjoint training set of many samples, over multiple epochs. 
The LTM learns incrementally about common concepts that can be used compositionally to represent unseen classes.

\noindent \textbf{Stage 2: Evaluate LTM+STM}: An evaluation has two steps performed in succession - STM training (encoding) and STM testing (inference). During both steps, LTM is used in inference mode (no learning occurs). STM is reset after each evaluation. 
\begin{itemize}
	\item \textit{Train}: A small `study' set is presented once. STM modules are set to train mode to learn the samples. 
	\item \textit{Test}: A small `recall' set is presented, STM modules are in inference mode. For each `recall' sample, the system is expected to retrieve the corresponding sample from the `study' set. If correct, it is considered to be `recognised' - an AHA moment! 
\end{itemize}

Training and Testing in STM occur rapidly over multiple internal cycles within one exposure to an external stimulus.

\subsection{AHA Model}
\label{sec:aha}

In this section we describe our implementation of the biological model given in Section~\ref{sec:comp_model}, named AHA - Artificial Hippocampal Algorithm. There are sub-sections for each subfield functional component as well the AHA specific training and testing procedures.
For the remainder of the paper we adopt a Functional(Biological) notation to clearly identify the functional instantiations and their biological counterparts.

AHA performs the role of the fast learning STM in Figure~\ref{fig:complementary_systems}. 
Since EC is considered to be a statistical learner, for simplicity it has been bundled into a unitary LTM representing a combined neocortex/EC.
LTM comprises a simple Vision Component, \VC, suitable for visual processing of image data.
It is trained to recognise common, simple visual features of the dataset, which are combined in the evaluation phase on unseen data. 

\begin{figure}[ht]
  \centering
    \includegraphics[width=0.98\columnwidth]{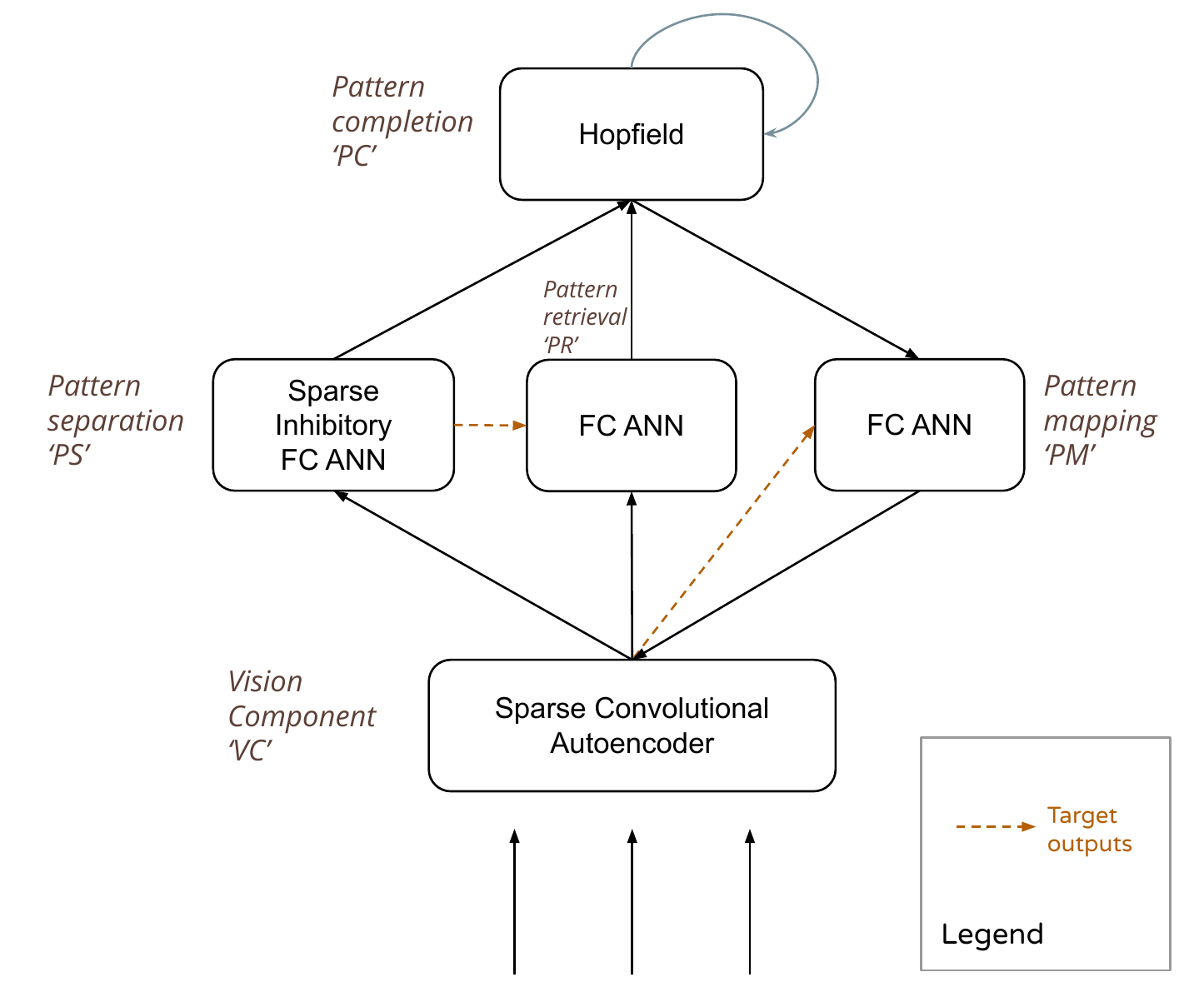}
  \caption{\textbf{AHA implementation mapping to biological subfields.} FC-ANN is an abbreviation of Fully-connected Artificial Neural Network. When the output of an ANN layer is used as the supervised target output for the training of another component, dashed arrows are used. All components are shallow and trained with local and immediate credit assignment.}
  \label{fig:aha}
\end{figure}

Figure~\ref{fig:aha} shows the components and connectivity of AHA used to implement the biological model, Figure~\ref{fig:biological_fields}. A detailed version is given in Appendix~\ref{app:app_system}.

\subsubsection{\PS\ - Pattern Separation}
\PS\ produces the increasingly sparse and orthogonal representations observed in the EC-DG-CA3 pathway. 
The idealised function of \PS\ is similar to a hashing function: similar inputs should produce dissimilar outputs.
It is achieved with a randomly initialised and fixed single-layer Fully-connected ANN with sparsity constraints and temporal inhibition.

Sparsity is implemented as a `top-$k$' ranking per sample, mimicking a local competitive process via inhibitory interneurons. Smaller values for $k$ are enough to produce less overlapping outputs, but orthogonality is further improved by replicating the sparse connectivity observed between DG-CA3 (as discussed in Section~\ref{sec:comp_model}). A portion, $\upsilon$, of the incoming connections are removed by setting weights to zero (similar to the sparsening technique of \cite{Ahmad2019}). 
Additionally, after a neuron fires (i.e. it is amongst the top-$k$), it is temporarily inhibited, mimicking the refractory period observed in biological neurons.

The first step of the temporal inhibition is to calculate the weighted sum $z_{i}$ for each neuron $i$. The inhibition term $\phi_{i}$ is applied with an elementwise multiplication and then a mask, $M$ is calculated for the top-$k$ elements.
We use an operator $\topk(a, b)$ that returns a `1' for the top $b$ elements in argument $a$, and `0' otherwise. 
	\begin{gather*}
		M = \topk(\phi \cdot z, k)
	\end{gather*}
The mask is applied with an elementwise multiplication to select the `active' cells that have a non-zero output.
	\begin{gather*}
		y = M \cdot z
	\end{gather*}
Inhibition decays exponentially with the hyperparameter $0 \leq \gamma \leq 1$, which determines the inhibition decay rate.
	\begin{gather*}
		\phi_{i}(t+1) = \phi_{i}(t) \cdot \gamma
	\end{gather*}
\PS\ is initialised with (uniformly distributed) random weights and does not undergo any training, similar to an Extreme Learning Machine \citep{Huang2006}.

As mentioned, the idealised function is similar to a hashing function. There have been other explorations of hashing with ANNs, such as chaotic neural networks \citep{Lian2007, Li2011}. However, they tend to be complex, multilayered and do not fit our biological plausibility constraints.
The \PS\ version may be useful in other contexts where orthogonality or pseudo-random non-clashing outputs is required.

\PS\ has 225 units, with a sparsity of 10 active units at a time. This and the other hyperparameters (described above) were chosen empirically to minimise resources and achieve orthogonal outputs on all images in an experimental run (see Section~\ref{sec:experimental_method}).

\subsubsection{\PC\ - Pattern Completion}
\PC\ comprises the recurrent auto-associative memory of CA3.
\PC\ is implemented with a Hopfield network, a biologically-inspired auto-associative content-addressable memory \citep{Hopfield1982, Hopfield1984}. It can store multiple patterns and recall crisp and complete versions of them from partial cues.

\PC's role is dedicated to storage and auto-associative recall, and does not perform pattern recognition. As such, \PC\ has the same number of neurons (225) as \PS\ with a fixed one-to-one connectivity between them, leaving the \PS\ to encapsulate biological connectivity between DG and CA3.

Unlike a standard Hopfield network, there are separate pathways for encoding and recall. \PS\ patterns are encoded by learning the appropriate recurrent weights for a given input. The recall cue is provided via the \PR\ network, with full explanation given in that sub-section below. 

We use graded neurons with input magnitude in the range $[-1, 1]$ and a $\tanh$ activation function. A gain $\lambda=2.7$ was empirically determined to be effective. A small portion $n=20$ of all neurons are updated at each step (of $N=70$ iterations), which speeds up convergence without any practical consequences (i.e. the slower case of $1$ neuron per step guarantees convergence). The Pseudoinverse learning rule \citep{Pal1996} was used to learn the feedback recurrent weights. It was chosen for convenience in that it increases capacity and is calculated in one time-step. Weights can also be learnt with more biologically realistic alternatives such as the original Hebbian learning or the Storkey learning rule \citep{Storkey1997}. The inputs to \PC\ are conditioned to optimise memoristion and retrieval from the Hopfield, detailed in Appendix~\ref{sec:app_pc}.

\subsubsection{\PR\ - Pattern Retrieval}
\label{sec:pr}
\PR\ models the connectivity between EC and CA3. Its role is to provide a cue to the \PC.
It is implemented with a 2-layer Fully-connected Artificial Neural Network (FC-ANN).  

\PC\ stores patterns from \PS\ and must be able to recall the appropriate pattern given a corresponding \VC\ input which is unlikely to be exactly the same or an exact subset of the memorised input (without the use of synthetic data). 
If \PS\ functions as intended, then small changes in \VC\ produce output from \PS\ that is orthogonal to the memorised pattern. This will not provide a meaningful cue to \PC; hence the role and importance of EC-CA3 connectivity. As explained in Section~\ref{sec:comp_model}, EC-CA3 connectivity constitutes a pattern recognition network that allows exploitation of the overlapping representation in EC that contains information about underlying concepts.
In effect, the sparsely activated \PC\ neurons form a `target' output for a given \VC\ pattern.
To train the pattern retrieval capability, the \VC\ is taken as input and the \PS\ output is used as internally generated labels for self-supervised training within the biological plausibility constraints. 
Once learned, the subsequent \PR\ outputs constitute a cue to recall a stored pattern in \PC.

In self-supervised learning \citep{Gidaris2019b}, usually prior knowledge about the task is used to set a pre-conceived goal such as rotation, with the motivation of learning generalisable representations. In the case of AHA, no prior is required. The motivation is separability. As such, the use of orthogonal patterns as labels, is very effective.

\PR\ is a 2-layer FC-ANN. 
Conceptually, the output layer represents the same entity as \PC, it is the same size (225 units) and the values are copied across.
2 layers were chosen to achieve the required complexity for the task within the biological constraints (Section~\ref{sec:bio_plaus}).
The hidden layer size is significantly larger (1000 units) to achieve the necessary capacity, empirically chosen to optimise test accuracy in the experimental conditions of training for many iterations on one batch (see Section~\ref{sec:experimental_method}).
Leaky-ReLU and Sigmoid activation functions are used in the hidden and output layers respectively. L2 regularisation is used to improve generalisation. 
Learning to retrieve the sparsely active target pattern is posed as multi-label classification with cross-entropy loss.

\subsubsection{\PM\ - Pattern Mapping}
\PM, the CA1 equivalent, maps from the highly orthogonal \PC\ representation, to the overlapping and distributed observational form.
\PM\ effectively learns to ground the orthogonal `symbolic' \PC\ representation for replay.
This emulates the way that during encoding, activated CA1 neurons form associative connections with the active EC neurons, achieved with self-supervised training, within the biological plausibility constraints.
A 2-layer FC-ANN is used with the output layer representing EC\textsubscript{out}. In this study we trained it to reconstruct the input images rather than the \VC\ output, making visualising the quality of the output intuitive.

A similar network was used for \PR\ to emulate connectivity between biological neurons of EC and CA3. 
In this case, it is being used to represent the subfield (CA1, a single layer of biological neurons) as well as the connectivity.
We postulate that only a very simple network is required due to the relatively limited scope of the experimental task (encoding of 20 characters). 
In addition, the afferent projections from EC\textsubscript{in} have been ignored in our implementation.
Re-evaluation of these simplifications may be important for extending the model.

Compared to \PR, a significantly smaller capacity is required. The hidden layer is 100 neurons wide, the output layer is constrained to the size of \VC. Both layers use the Leaky-ReLU activation functions, training is conducted with MSE loss function and L2 regularisation. The hyperparameters are chosen empirically based on loss values and visual inspection of the quality of reconstructions.

\subsubsection{\VC\ - Vision Component}
The role of \VC\ is to process high dimensional sensor input and output relatively abstract visual features that can be used as primitives compositionally.
A single layer convolutional sparse autoencoder based on \citep{Makhzani2013,Makhzani2015} provides the required embedding (see Appendix~\ref{sec:app_scae} for details). However, in Omniglot there is a lot of empty background that is encoded with strong hidden layer activity. Lacking an attention mechanism, this detracts from compositionality of foreground features. To suppress encoding of the background, we added an `Interest Filter' mechanism.

The Interest Filter loosely mimics known retinal processing (see Appendix~\ref{sec:app_vc_if}).
The retina possesses centre-surround inhibitory and excitatory cells that can be well approximated with a Difference of Gaussians (DoG) kernel \citep{Enroth-Cugell1966, Young1987, Mcmahon2004}.

\subsubsection{Training and Testing}
In accord with the Training and Testing Framework (Section~\ref{sec:train_test_fwk}), the hippocampal components \PC, \PR\ and \PM\ train (encode) and test (retrieve) within one stimulus exposure each. The learning process during that exposure is implemented as $N=60$ mini-batches with constant input. During retrieval, the \PC\ converges over $70$ recursive iterations.
\PS\ does not undergo training.
All the components are reset between evaluation steps, meaning the optimizer state (e.g. momentum) and all learned parameters are lost and replaced with re-initialised values.

The use of separate train and test phases is modelled after hippocampal encode and retrieval phases governed by theta rhythms (Section~\ref{sec:comp_model}).
Resetting between experiments is consistent with the way other hippocampal studies were conducted (discussed in Section~\ref{sec:comp_model}) and supported by evidence that during retrieval, there is depotentiation in synapses that allows rapid forgetting \citep{Hasselmo2002}.

\subsection{Theory of Operation}
\label{sec:theory_of_operation}

\subsubsection{From Sensor Observations to Symbols}
As discussed above for the biological hippocampus, CA3 patterns are highly orthogonal, effectively random encodings of the EC engram, thereby accomplishing pattern separation. 
They are stable, distinct and statistically unrelated to other even similar observations, therefore we hypothesise that they fulfil the role of a symbolic representation.
Taken as a whole, the EC-CA3 network maps from observations grounded in perceptual processing to symbols, and CA1 performs the inverse mapping. 

This characteristic raises the possibility of symbolic reasoning on grounded sensor observations. Recent work \citep{Higgins2018} shows the possibility of using symbols that represent generative factors, in novel combinations, to ``break away from'' the training distribution. In the cited work the symbols are `labels' provided externally. However, biological `agents' only receive information through sensors, even when it is symbolic in nature, such as spoken or written words. This could be supported by a hippocampal architecture, in which the symbols are self generated.

\subsubsection{Collaboration of \PS\ and \PR}
\label{sec:theory_of_operation_collab}
A core component of operation is the manner in which the pattern separation pathway, \PS, and the pattern retrieval pathway \PR, are unified.
\PS\ sets a target for both \PR\ and \PC\ to learn, providing a common representational `space'.
This makes it possible to separate encoding and retrieval (to and from \PC), between the separation and completion pathways respectively.
In this way, they don't dilute each other, but operate to their strengths.

The fact that \PS\ provides a strong symbolic-style target for \PR\ to learn, makes learning that target with self-supervised learning, very effective.
In addition, orthogonality between samples is preferable for efficacy of the \PC\ Hopfield network \citep{Hopfield1982}.
In turn, the high quality orthogonal outputs from \PC\ allow \PM\ to learn to reconstruct effectively.

\subsubsection{Unifying Separation and Completion}
\label{sec:theory_of_operation_sep_comp}
Both separation and completion are necessary for the range of experiments reported here. Separation allows storage of a unique form of the memory, and completion recalls the most `similar' memorised form. It does not matter whether observational variation is caused by occlusion, noise or different exemplars of the same class.
Separation and completion are conflicting capabilities requiring separate pathways.
Unification is achieved through collaboration of \PS\ and \PR\ described above.

\subsubsection{Continual Learning of More Abstract Concepts}
The hippocampus' role in replay and consolidation, facilitating continual learning without catastrophic forgetting, is recognised as a valuable ability for intelligent machines \citep{Kumaran2016}, particularly where there are few experiences of new tasks.
A hippocampal architecture that functions like AHA may provide an additional aspect of continual learning.
As discussed, the STM (hippocampus) receives compositional primitive concepts from LTM (neocortex). New conjunctions of concepts are learnt in STM as composite, more abstract, concepts. Following consolidation to LTM, these new concepts may in turn serve as primitives for subsequent stages of learning. We hypothesise that this could confer an ability to build progressively more abstract concepts hierarchically.

\subsection{Baseline STM Model - \FNN}
\label{sec:comparison_model}
In this section, we describe an alternative STM for comparison with AHA.
The objective is to test AHA against a standard ANN architecture optimised for the same tasks.
It must also learn fast given only one exposure, so it is referred to as \FNN.
Like AHA, \FNN\ reconstructs a memorised input based on an internally generated cue, and is therefore a form of auto-associative memory. 
Again like AHA, the input and output data structures are not identical. \VC\ provides the input, and the output are reconstructed images (this is for ease of analysis rather than biological compatibility).

\FNN\ is a 2-layer Fully-connected ANN. 
We empirically optimised the learning rate, number and size of hidden layers, regularisation method and activation functions of the model.
The resulting model has a hidden layer with 100 units, L2 regularisation and Leaky-ReLU activation function. The output layer size is constrained to the size of the output image.
As \FNN\ reconstructs an input signal, the first layer acts as an encoder that produces a latent representation and the second layer a decoder that maps this back to the image space.

\section{Method}
\label{sec:experimental_method}
The experiments test the ability to recognise the input in response to a cue (retrieve the correct memory internally), and to replay it in the form of the originating input. 
A variety of experimental methods are reported in the literature for other CLS-style models \citep{Rolls1995, Norman2003, Greene2013, Ketz2013, Schapiro2017a}. 
We chose a test that has become standard in the ML literature for \oneshot\ learning.
Our intention is to use a method that fits the existing style of tests and that we can use to compare performance to other non-hippocampal state-of-the-art methods.

The experiments test the ability to learn from a single exposure and to retrieve appropriate memories (for replay to LTM) in the context of \oneshot\ classification tasks.
Performance is assessed with quantitative and qualitative results.
The ability to recognise is measured with classification accuracy, and the quality of end-to-end retrieval of appropriate memories is assessed with recall-loss and through visual inspection of signals as they propagate through the respective architectures.

We test the performance of the LTM on its own, which we refer to as the baseline, and compare it to architectures that combine an LTM and STM. The two combined architectures are AHA (LTM+AHA) and \FNN\ (LTM+\FNN).
For all architectures, classification is performed by comparing internal states between memorisation and retrieval patterns.
The architectures are not explicitly trained to classify. 
In the case of AHA, we used internal states at different points in the architecture to show the varied performance of \PR\ and \PC.
Conventional ablation is not possible because \PC\ is fully dependent on the earlier stages \PS\ for encoding and \PR\ for retrieval.

The experiments are based on the \oneshotGen\ task from \citep{Lake2015}, which uses the Omniglot dataset of handwritten characters from numerous alphabets.
It was chosen because it has become an accepted benchmark for \oneshotGen\ learning with many published results (reviewed in \citep{Lake2019}. 
We extended it to test robustness to more realistic conditions and a wider range of capabilities.

The following sections detail the original benchmark, followed by our extended version, conditions common to all tests and details of how performance was evaluated.

\subsection{Omniglot Benchmark}
Using the terminology from \citep{Lake2015}, we refer to this task as \oneshotGen. It involves matching distinct exemplars of the same character class, without labels, amongst a set of distractors from other character classes. Strong generalisation is required to complete the task, as well as some pattern separation to distinguish similar character classes.

Referring to the train/test framework in Section~\ref{sec:train_test_fwk}, the system is pre-trained with a `background' set of 30 alphabets.
Then in the core task, using 10 alphabets from a separate `evaluation' set, a single `train' character is presented. The task is to identify the matching character from a `test' set of 20 distinct characters drawn from the same alphabet and written by a different person.
This is repeated for all 20 characters to comprise a single `run'. The experiment consists of 20 runs in total (400 classifications).
Accuracy is averaged over the 20 runs.\footnote{Note that we used the approach in \citep{Lake2015}, calculated as average of the average of each run.}
The method used to determine the matching character varies between studies. In this work, we use minimum MSE of a given internal state (the chosen state depends on which model is being assessed).\footnote{Note that the MSE calculation is straightforward as sparse vectors are not index based, but are implemented as dense vector that are sparsely activated.}

The characters and alphabets were originally selected to maximise difficulty through confusion of similar characters.
In addition, the test condition of within-alphabet classification is more challenging than between-alphabet.
Some other \oneshotGen\ studies that base their experiments on the Omniglot benchmark used only 5 alphabets, tested between-alphabet classification, and used augmented datasets, making the problem significantly easier \citep{Lake2019}.

\subsection{Omniglot Extended Benchmark}
Instance learning enables agents to reason about specifics rather than class categories e.g. an ability to recognise your own specific cup. 
The episodic hippocampal models referenced in Section~\ref{sec:comp_model} have traditionally focused on this type of task.
Therefore, we extended the Omniglot classification benchmark to include instance learning.

We refer to this task as \oneshotInst. It requires strong pattern separation, as well as some generalisation for robustness. It is the same as \oneshotGen, except that the train character exemplar must be matched with the exact same exemplar amongst a set of 20 distractor exemplars of the \emph{same} character class. 
Being the same character class, all the exemplars are very similar making separation difficult.
In each run, the character class and exemplars are selected by randomly sampling without repeats from the `evaluation' set.

In addition, we explored robustness by introducing image corruption to emulate realistic challenges in visual processing that could also apply to other sensory modalities.
Noise emulates imperfect sensory processing. For example, in visual recognition, the target object might be dirty or lighting conditions changed. Occlusion emulates incomplete sensing e.g. due to obstruction by another object. Robust performance is a feature of animal-like learning that would confer practical benefits to machines, and is therefore important to explore \citep{Ahmad2019}.
Occlusion is achieved with randomly placed circles, completely contained within the image. 
Noise is introduced by replacing a proportion of the pixels with a random intensity value drawn from a uniform distribution.

For both tests, instead of presenting 1 test character at a time, all 20 are presented simultaneously.
This is compatible with other hippocampal studies referenced above and requires a short term memory of moderate capacity, which the hippocampal model provides. Accuracy and recall-loss are measured for increasing noise and occlusion from none, to almost complete corruption, in 10 increments. The highest level is capped at just below $98\%$ corruption, to ensure some meaningful output. 
Every test is repeated with 10 random seeds.

The full set of experiments are summarised in the table below:\newline\newline

\newcolumntype{L}[1]{>{\raggedright}p{#1\linewidth}}
\noindent\begin{tabular}{L{0.2}|L{0.2}|L{0.22}|L{0.2}}

	{\small \textbf{Experiment}} &
	{\small \textbf{Match}} &
	{\small \textbf{Distractors}} &
	{\small \textbf{Primarily Testing}} \tabularnewline \hline

	{\small \OneshotGen} &
	{\small Character class} &
	{\small Exemplars from other classes} &
	{\small Generalisation} \tabularnewline \hline

	{\small \OneshotInst} &
	{\small The exemplar of a class} &
	{\small Exemplars of same class} &
	{\small Separation} \tabularnewline
\end{tabular} \\

\subsection{Performance Analysis}
The accuracy calculation is described in the detail of the Omniglot Benchmark experiment above.
`Recall-loss' is the MSE difference between the original train image and a recalled image in grounded, observational form. 
Qualitative results are analysed through visual inspection.

\section{Results}
In this section we present the experimental results organised primarily by evaluation methods: accuracy, recall-loss and qualitative analysis.
All experiments were conducted over $10$ random seeds. The plots for accuracy and recall-loss show the mean value in bold with medium shading for one standard deviation, and lightly shaded portion demarcating minimum and maximum values.

\subsection{Accuracy}

\begin{figure*}
     \centering
     \begin{subfigure}{0.48\textwidth}
         \centering
	 	 \includegraphics[width=\textwidth]{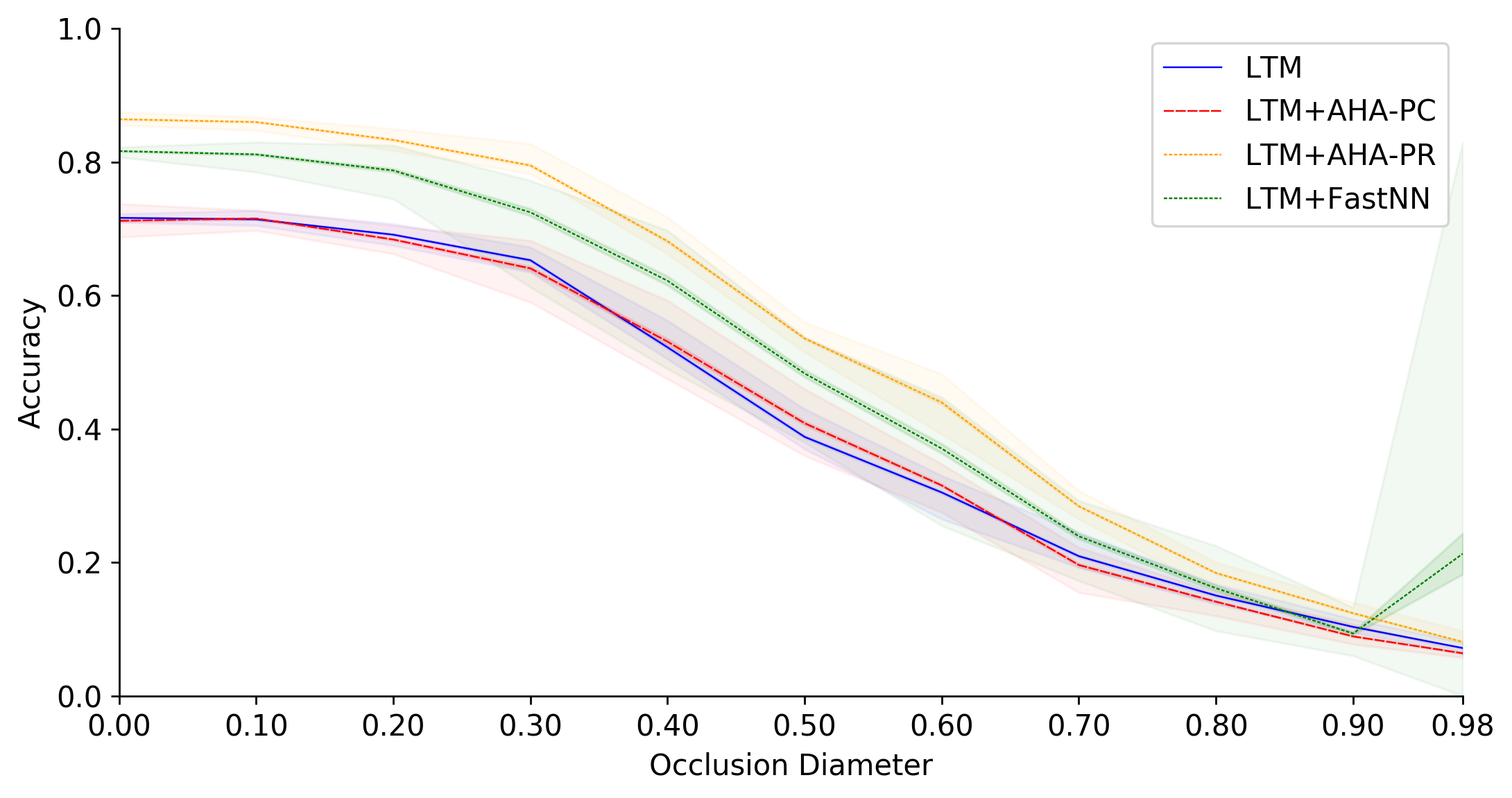}
		 \caption{\OneshotGen\ with Occlusion}
		 \label{fig:accuracy_plots_class_occ}
     \end{subfigure}
     \hfill
     \begin{subfigure}{0.48\textwidth}
         \centering
		 \includegraphics[width=\textwidth]{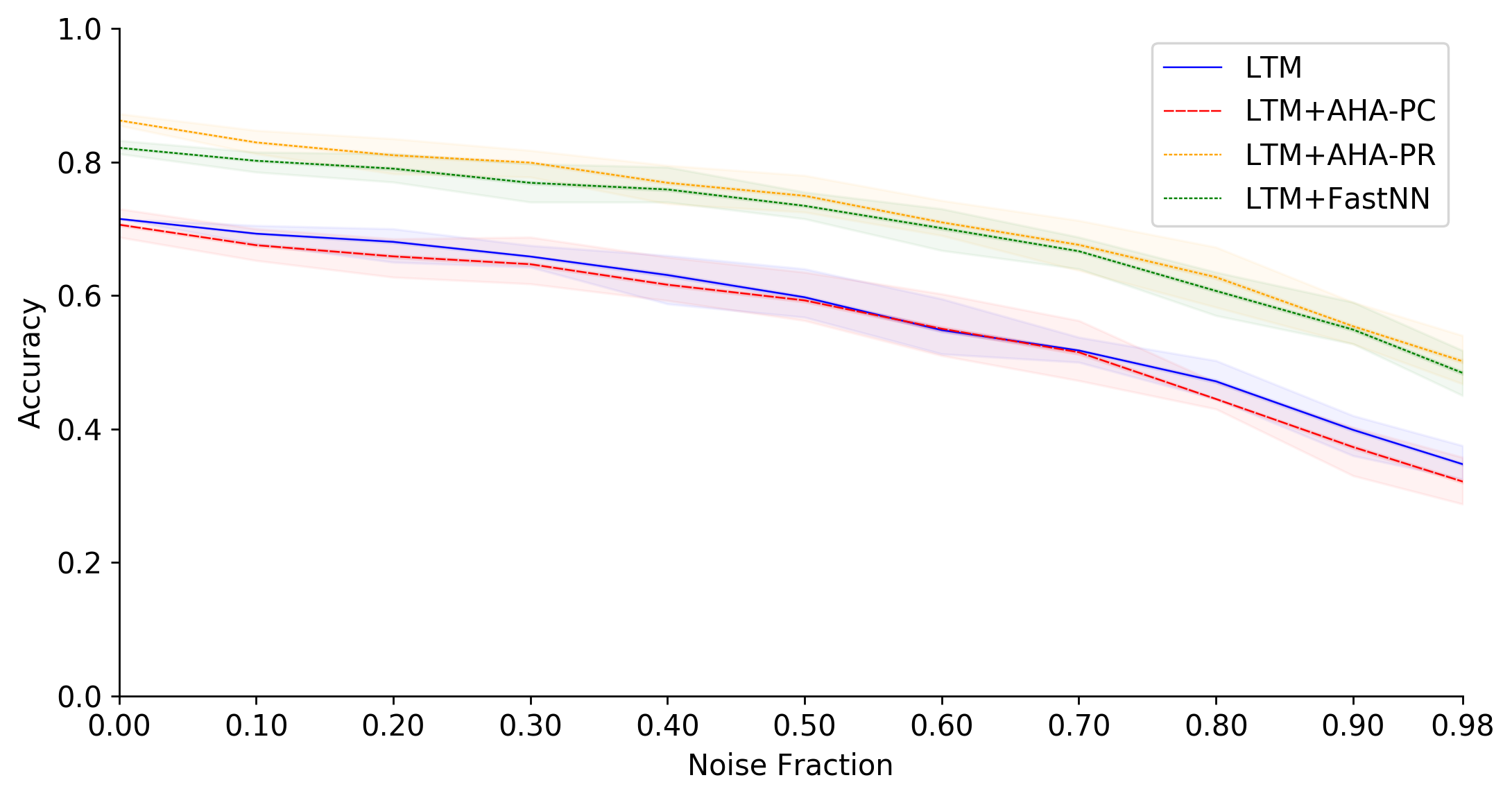}
  		 \caption{\OneshotGen\ with Noise}
  		 \label{fig:accuracy_plots_class_noise}
     \end{subfigure}

     \begin{subfigure}{0.48\textwidth}
		  \centering
		  \includegraphics[width=\textwidth]{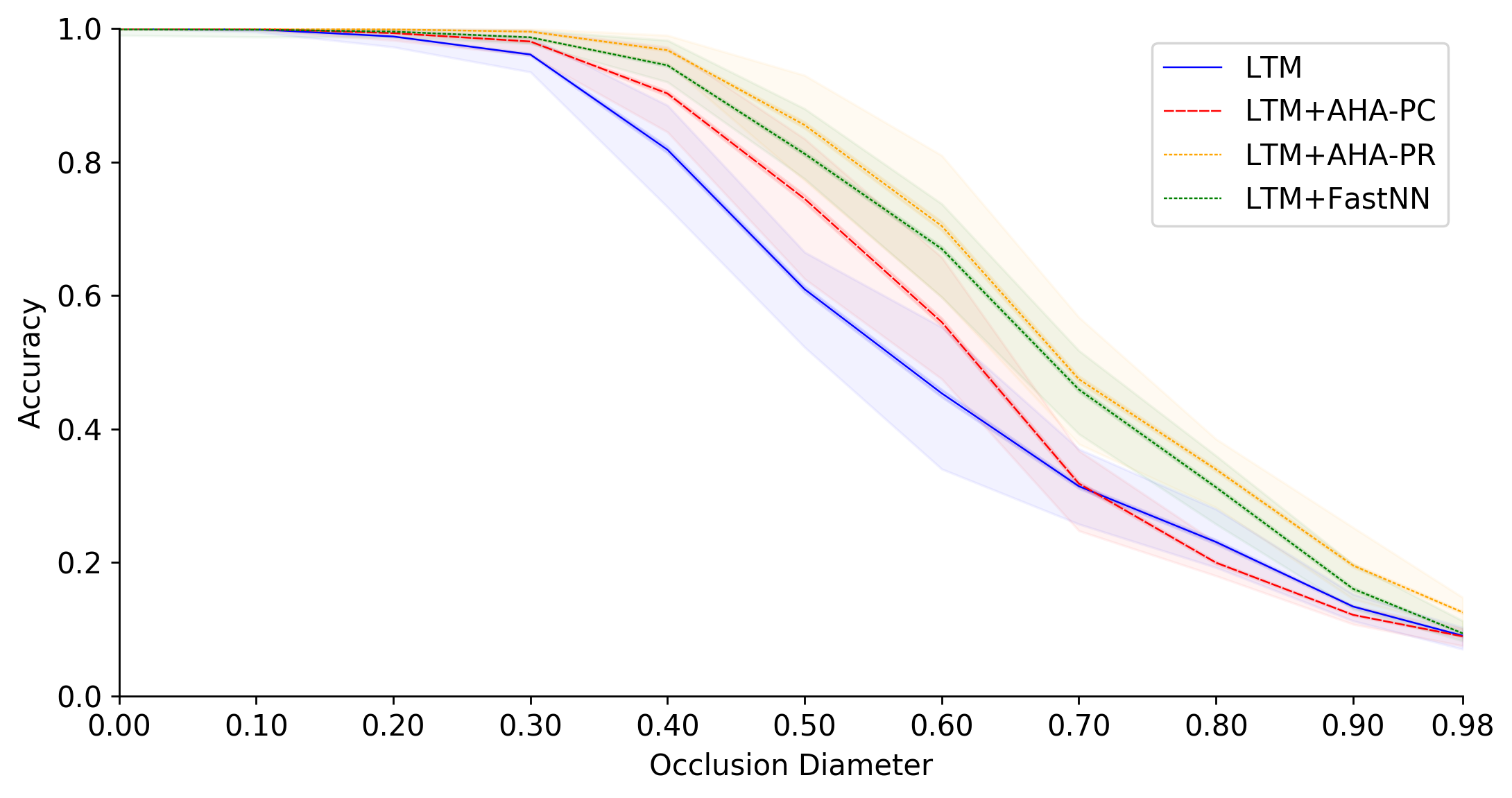}
		  \caption{\OneshotInst\ with occlusion}
		  \label{fig:accuracy_plots_inst_occ}
	 \end{subfigure}	
     \hfill
     \begin{subfigure}{0.48\textwidth}  
		  \centering
	      \includegraphics[width=\textwidth]{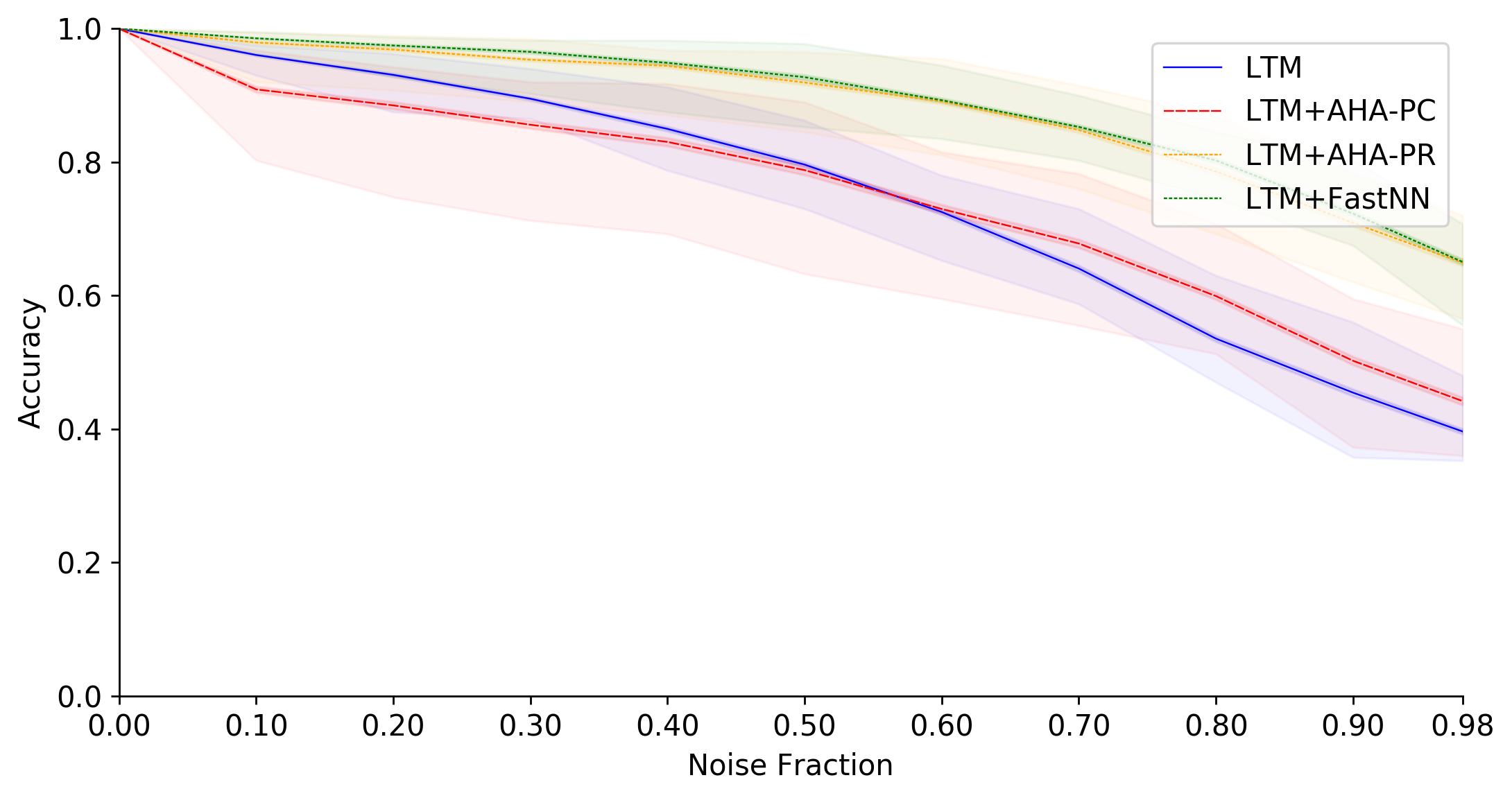}
	      \caption{\OneshotInst\ with noise}
	      \label{fig:accuracy_plots_inst_noise}
	 \end{subfigure}
     \caption{\textbf{\OneshotInst\ accuracy vs occlusion and noise.} LTM+AHA and LTM+\FNN\ improve performance over the baseline (LTM) for both types of image corruption. LTM+AHA has an advantage over LTM+\FNN\ for occlusion, minor for noise. Occlusion diameter as a fraction of image side length, noise is expressed as a fraction of image area.}
     \label{fig:accuracy_plots}
\end{figure*}

\subsubsection{\OneshotGen\ Task}
Accuracy results of the \oneshotGen\ task are shown in Figures \ref{fig:accuracy_plots_class_occ} and \ref{fig:accuracy_plots_class_noise}.
LTM on its own comprises the baseline. Without image corruption, classification accuracy is $71.6\%$. 
Moderate occlusion has a more damaging impact than noise, until very high levels where the character is mostly covered, leading to a plateau (overall an inverse sigmoid profile).
Noise effects all features equally and gradually, whereas occlusion increases the likelihood of suddenly removing important topological features.
\FNN\ and AHA follow the same overall trends as LTM.

The AHA pattern retrieval network \PR\ boosts performance over the baseline significantly by almost $15\%$ to $86.4\%$ at no noise or occlusion.
\PR\ has a strong advantage over baseline for all conditions.
As extreme occlusion begins to cover most of the character, the accuracy of \PR, \PC\ and LTM converge. However, the advantage is maintained over all noise levels, where there is still a signal at high levels.
For all conditions, \PC\ classification accuracy is no better than LTM.

\FNN\ improves on the baseline accuracy by $10.3\%$, $4.4\%$ less than the LTM+AHA improvement. 
The advantage of LTM+AHA over LTM+\FNN\ is significant over almost all levels of occlusion. The advantage is minor for all levels of noise.

For context, reported accuracy without noise or occlusion is contrasted with other works in Table~\ref{table:comparison}. Existing values are reproduced from \citet{Lake2019}.

\begin{table}[ht]
\begin{center}
\begin{tabular}{ l l c }
\hline
\textbf{Algorithm} & \textbf{Citation} & \textbf{Accuracy \%} \\ \hline
\textit{Human} & \citep{Lake2019} & \textit{95.5} \\ \hline
BPL & \citep{Lake2019} & 96.7 \\ \hline
RCN & \citep{George2017} & 92.7 \\ \hline
Simple Conv Net & \citep{Lake2019} & 86.5 \\ \hline
\textbf{LTM+AHA} &  & \textbf{86.4} \\ \hline
Prototypical Net & \citep{Lake2019} & 86.3 \\ \hline
VHE & \citep{Hewitt2018} & 81.3 \\ \hline
\textbf{LTM+\FNN} &  & \textbf{81.9} \\ \hline
\end{tabular}
\caption[Table caption text]{\textbf{Comparison of algorithms for \oneshotGen, without image corruption.} LTM+AHA is competitive with state-of-the-art convolutional approaches whilst demonstrating a wider range of capabilities.}
\label{table:comparison}
\end{center}
\end{table}

\subsubsection{\OneshotInst\ Task}
Accuracy results for the \oneshotInst\ task are shown in Figures \ref{fig:accuracy_plots_inst_occ} and \ref{fig:accuracy_plots_inst_noise}.
LTM Accuracy is perfect at low levels of image corruption, remaining almost perfect in the case of occlusion, until approximately one third of the image is effected. 
For all models, the same trends are observed as for the \oneshotGen\ task, with some salient features.

For AHA, \PR\ accuracy remains extremely high, close to $100\%$ until a $10\%$ greater level of occlusion than for LTM (i.e. addition of AHA increases tolerance to occlusion).
The advantage over LTM increases with increasing corruption, fading away for occlusion but continuing to grow for noise.
Note also that in the case of \oneshotInst\, which is the focus of most episodic work in computational models, unlike the \oneshotGen\ experiment, \PC\ confers a significant advantage that grows with occlusion until it converges as the task becomes too difficult to achieve at all. A possible explanation is that the cue provided to \PC\ is more likely to be closest to the correct attractor state.

LTM+\FNN\ also improves on the baseline. It has worse accuracy than LTM+AHA for a given level of occlusion (less substantial than \oneshotGen), and equal accuracy for varying levels of noise.

\subsection{Recall}
\begin{figure*}[ht]
     \centering
     \begin{subfigure}{0.48\textwidth}
         \centering
	 	 \includegraphics[width=\textwidth]{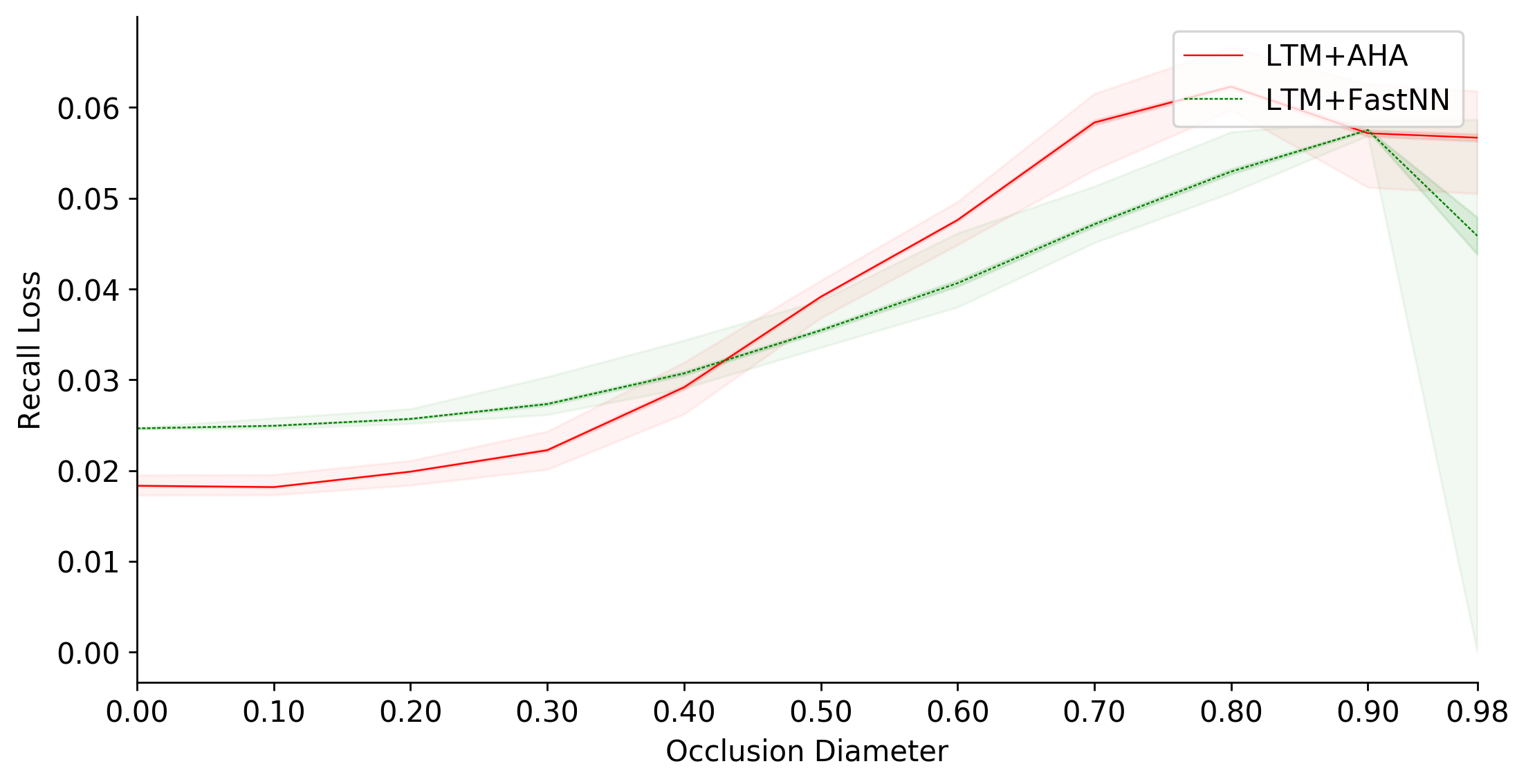}
		 \caption{\OneshotGen\ with occlusion}
     \end{subfigure}
     \hfill
     \begin{subfigure}{0.48\textwidth}
         \centering
		 \includegraphics[width=\textwidth]{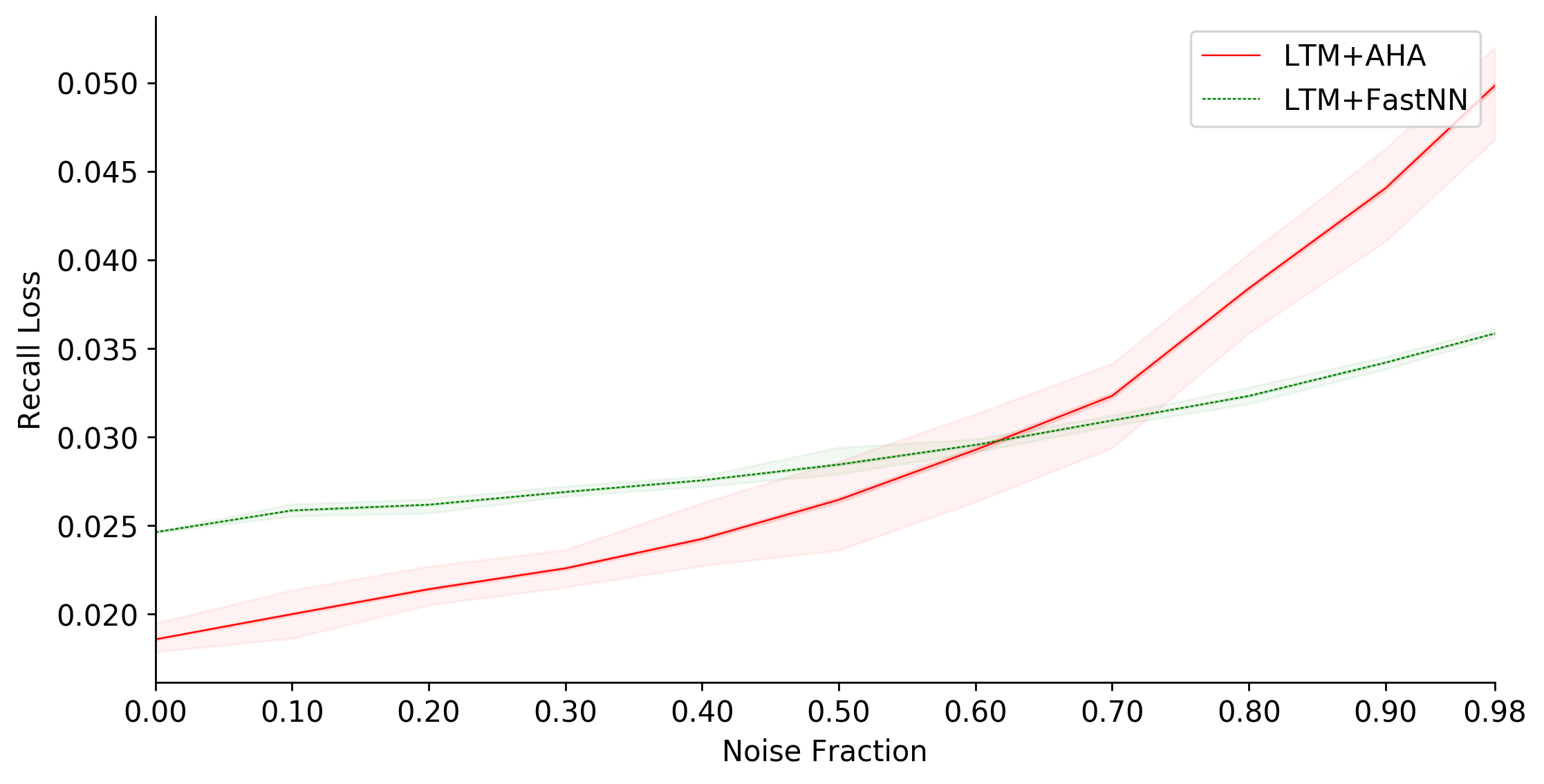}
  		 \caption{\OneshotGen\ with Noise}
     \end{subfigure}

     \begin{subfigure}{0.48\textwidth}     
		  \centering
		  \includegraphics[width=\textwidth]{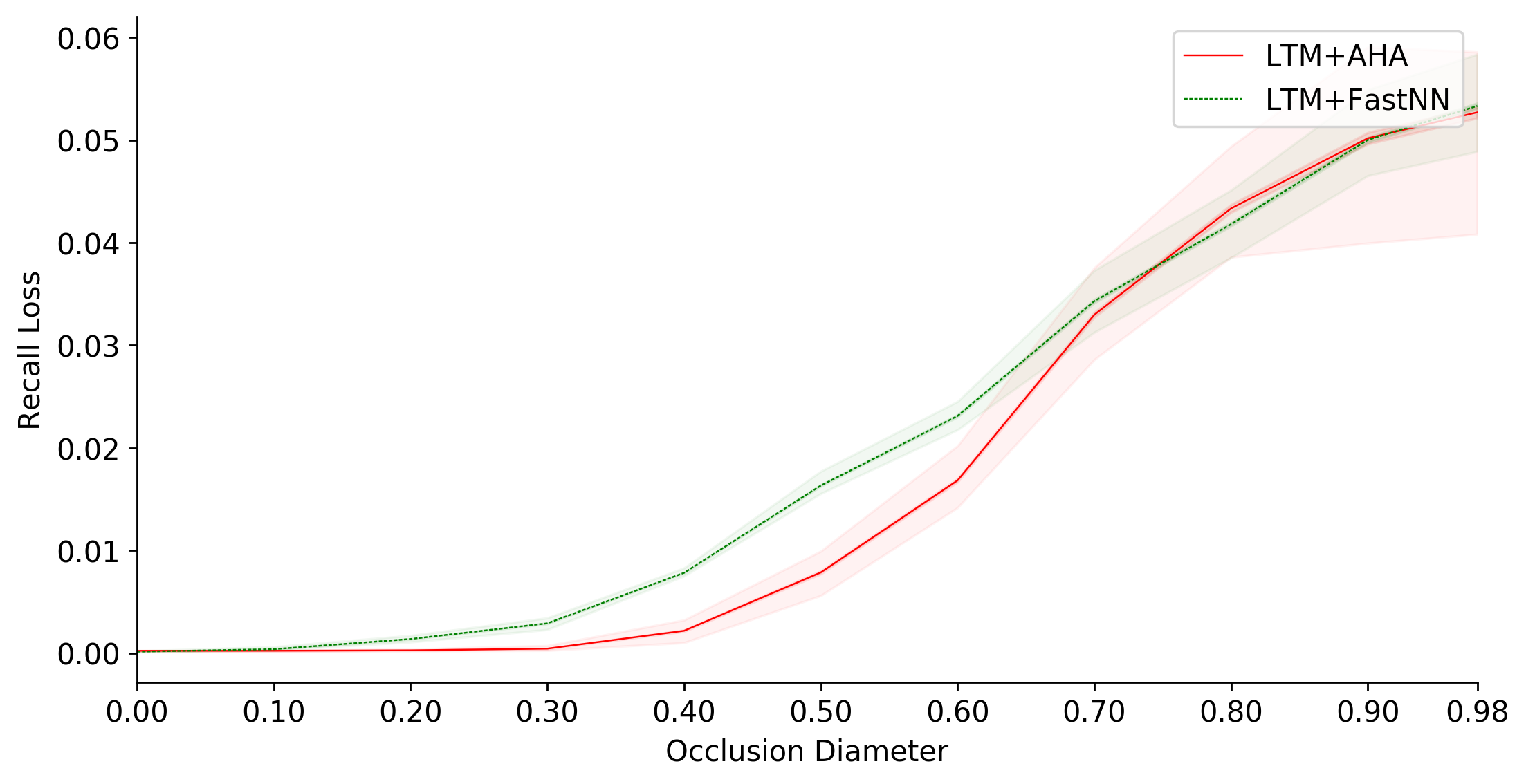}
		  \caption{\OneshotInst\ with occlusion}
	 \end{subfigure}	
     \hfill
     \begin{subfigure}{0.48\textwidth}  
		  \centering
	      \includegraphics[width=\textwidth]{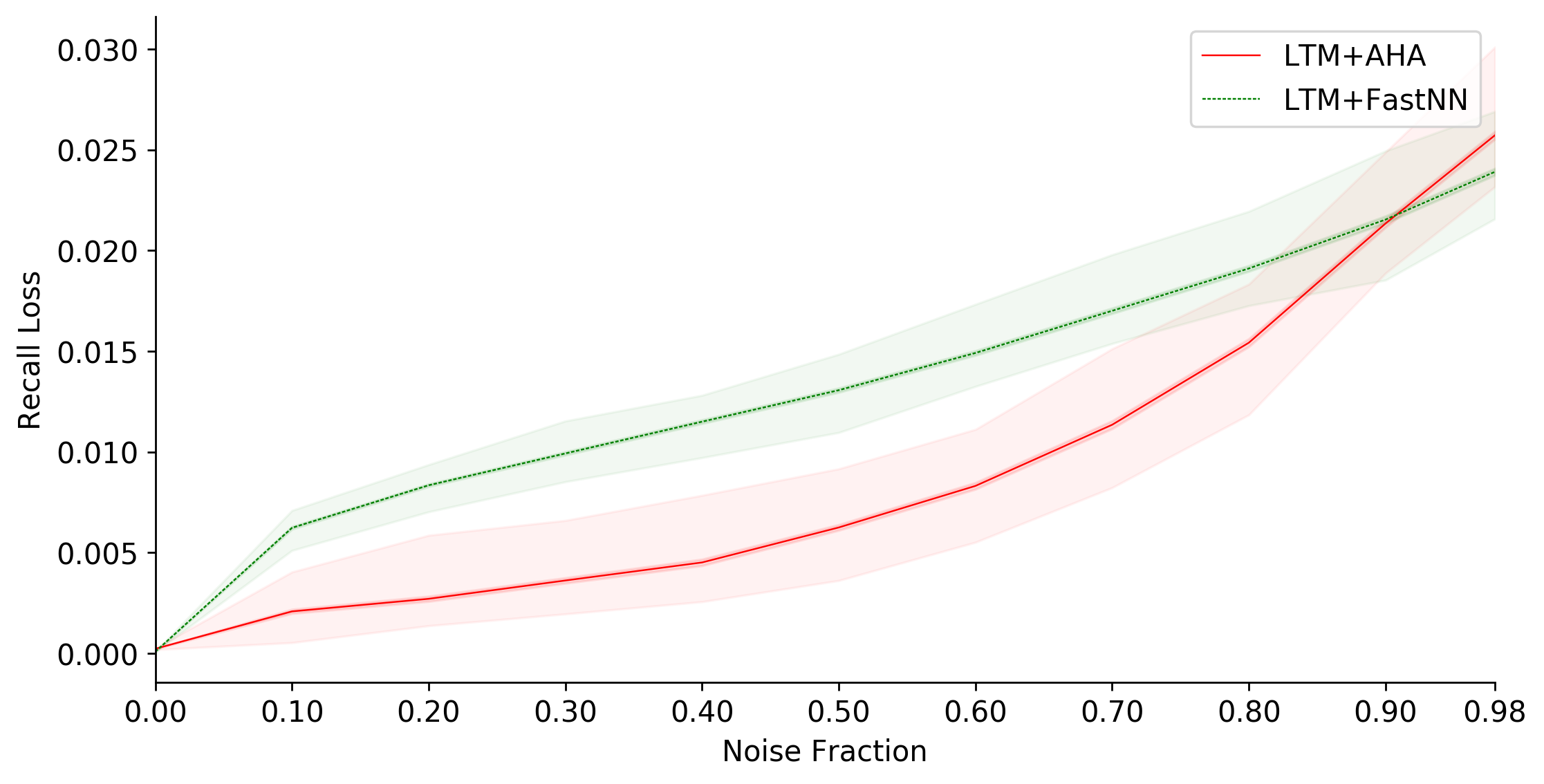}
	      \caption{\OneshotInst\ with noise}
	 \end{subfigure}
     \caption{\textbf{Recall-loss vs occlusion and noise.} The two STM models follow similar trends. AHA is superior given moderate to high image corruption, where it sometimes retrieves an accurate copy of the wrong character. Occlusion diameter is expressed as a fraction of image side length, noise as a fraction of image area.}
     \label{fig:replay_loss}
\end{figure*}

Recall-loss is plotted versus occlusion and noise for all experiments in Figure~\ref{fig:replay_loss}.
In the \oneshotGen\ experiment, LTM+AHA demonstrates better performance than LTM+\FNN\ under moderate occlusion and noise. At higher levels of corruption, LTM+AHA may retrieve a high quality image of the wrong character, which results in a higher loss than lower-quality images retrieved by LTM+\FNN.

In the \oneshotInst\ experiment, this character confusion is less likely to occur and LTM+AHA is superior or equal to LTM+\FNN\ under all meaningful levels of image corruption.

Both LTM+AHA and LTM+\FNN\ loss drops at extreme occlusion, which at first may seem counter-intuitive. 
With even a small portion of the character present in the image, the networks produce a more recognisable character output. This is more pronounced for AHA, which through the action of \PC\ converges on an attractor state. 
In comparison, when the character is completely occluded, the output is a faint blur consisting of a superposition of all of the learnt image (see figures in Appendix~\ref{app:high_corruption}).
In the former case, the loss is the difference between 2 strong characters (crisp in the case of AHA) - they differ most of the time at high levels of occlusion. In the latter case, it is the difference between 1 strong character and an effectively blank image, a lower loss even though it never recalls the correct image.

\subsection{Qualitative Analysis}
\label{sec:results_qa}
\begin{figure*}[ht]
	\centering
	\begin{subfigure}{\textwidth}
		\centering
		\includegraphics[width=\textwidth]{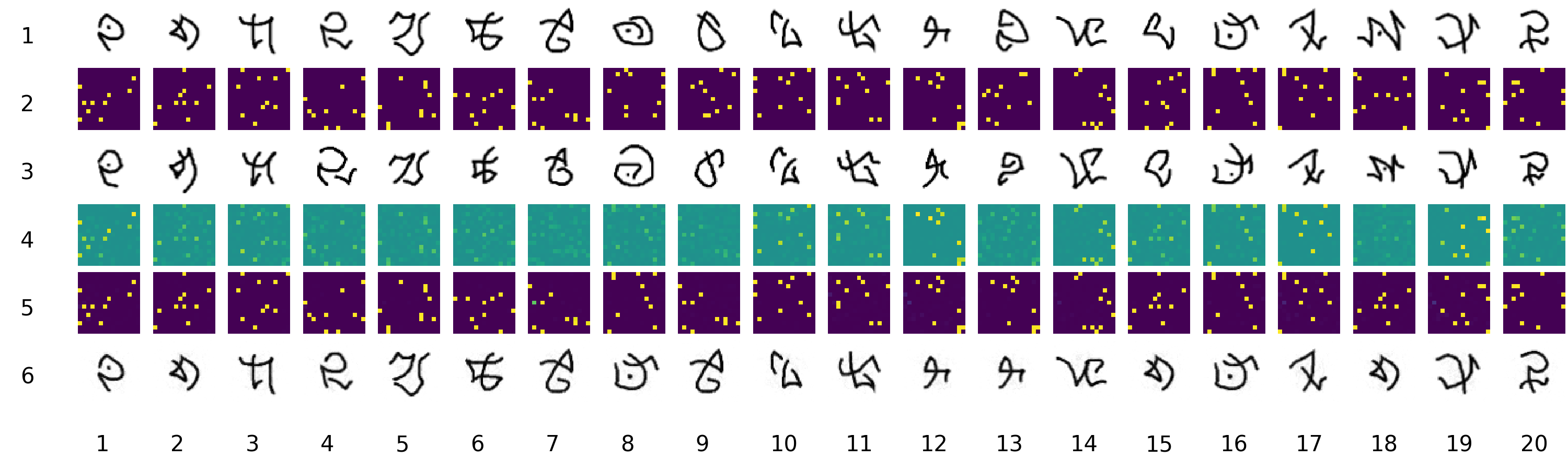}
		\caption{No occlusion or noise.}
		\label{fig:generalization_images}
	\end{subfigure}
	\bigskip
	\begin{subfigure}{\textwidth}
		\centering
		\includegraphics[width=\textwidth]{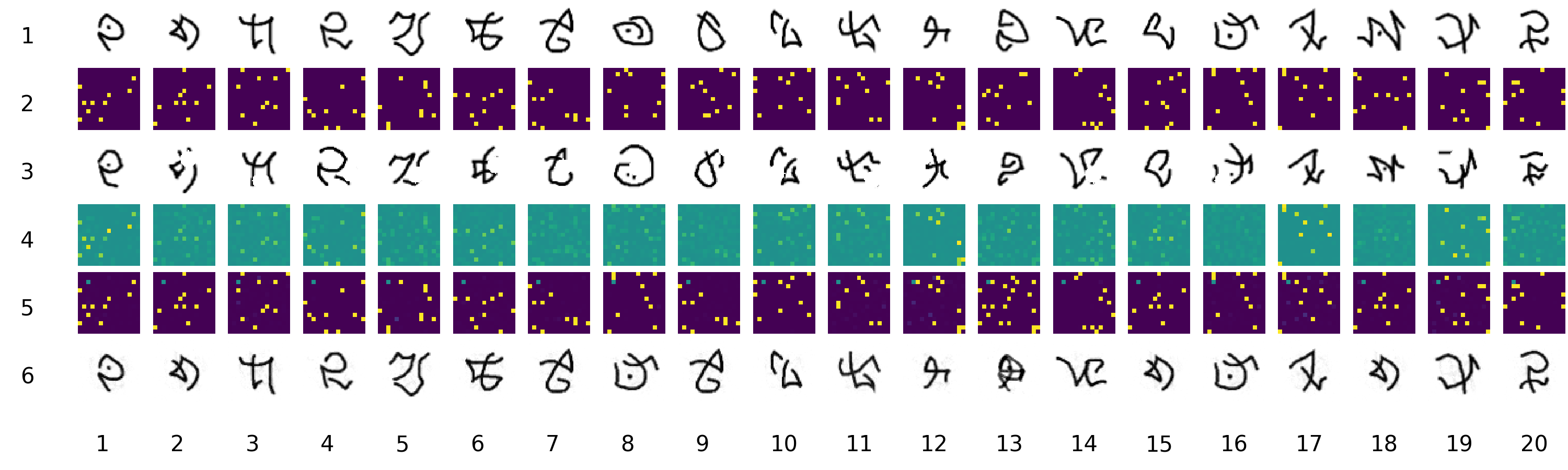}
		\caption{With occlusion (circle, diameter=0.3).}
		\label{fig:generalization_images_occl}
	\end{subfigure}
	\bigskip
	\begin{subfigure}{\textwidth}
		\centering
		\includegraphics[width=\textwidth]{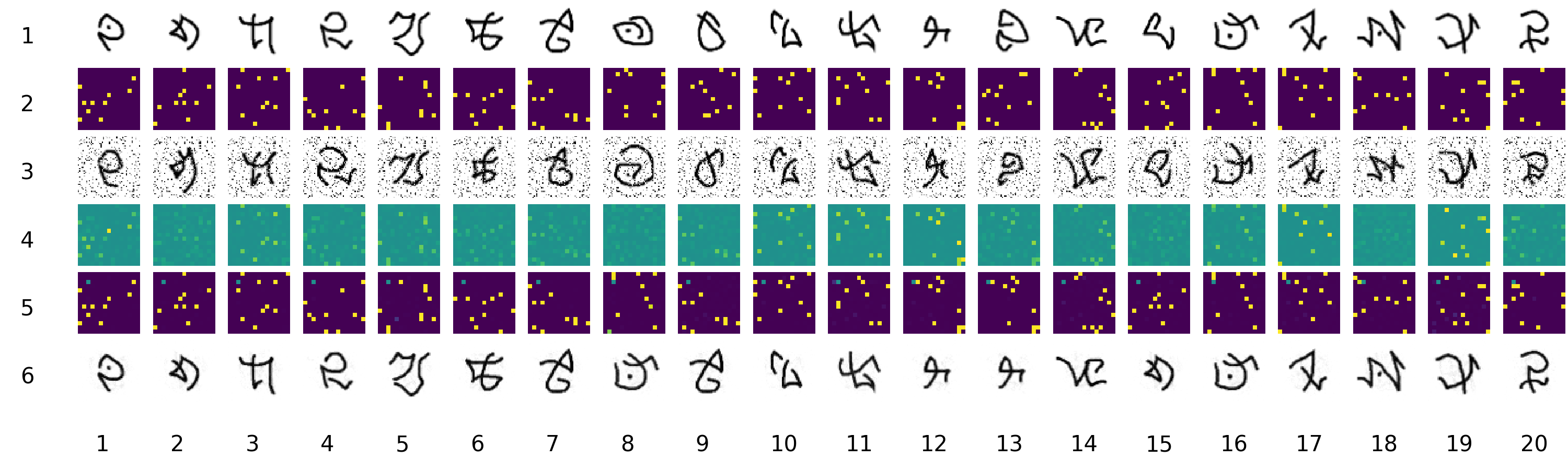}
		\caption{With noise (fraction=0.3).}
		\label{fig:generalization_images_noise_03}
	\end{subfigure}
	\caption{\textbf{\OneshotGen\ test patterns as they propagate through AHA}. See Table~\ref{table:legend} for legend.}
	\label{fig:class_qual}
\end{figure*}

\begin{figure*}[ht]
	\begin{subfigure}{\textwidth}
		\centering
		\includegraphics[width=\textwidth]{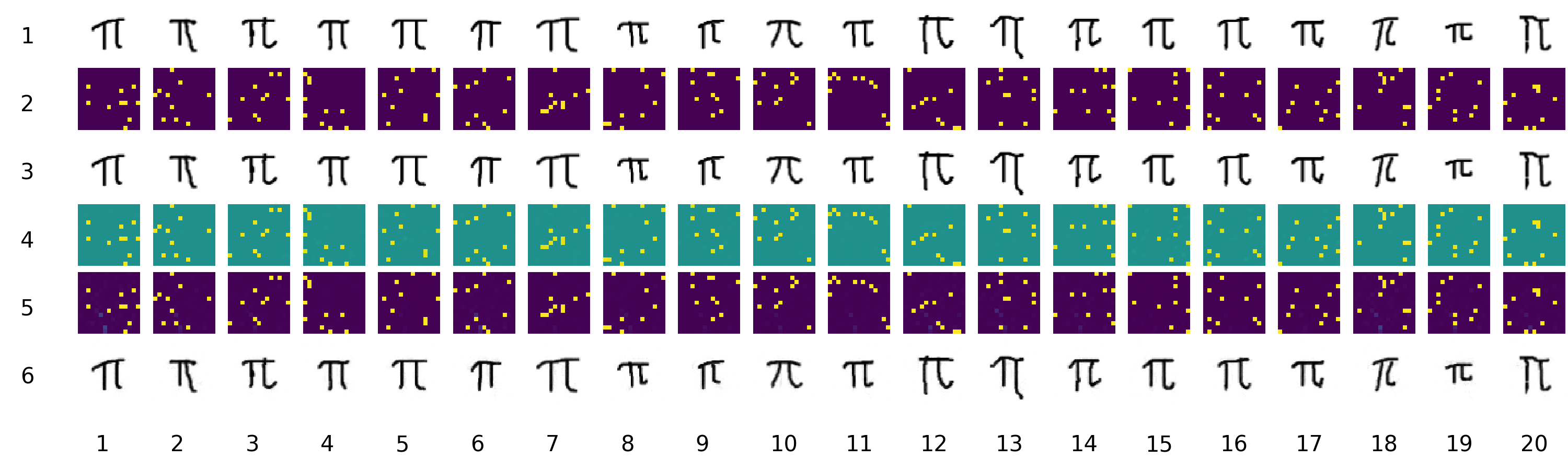}
	  \caption{No occlusion or noise.}\label{fig:instance_images}
	\end{subfigure}
	\bigskip
	\begin{subfigure}{\textwidth}
		\centering
    	\includegraphics[width=\textwidth]{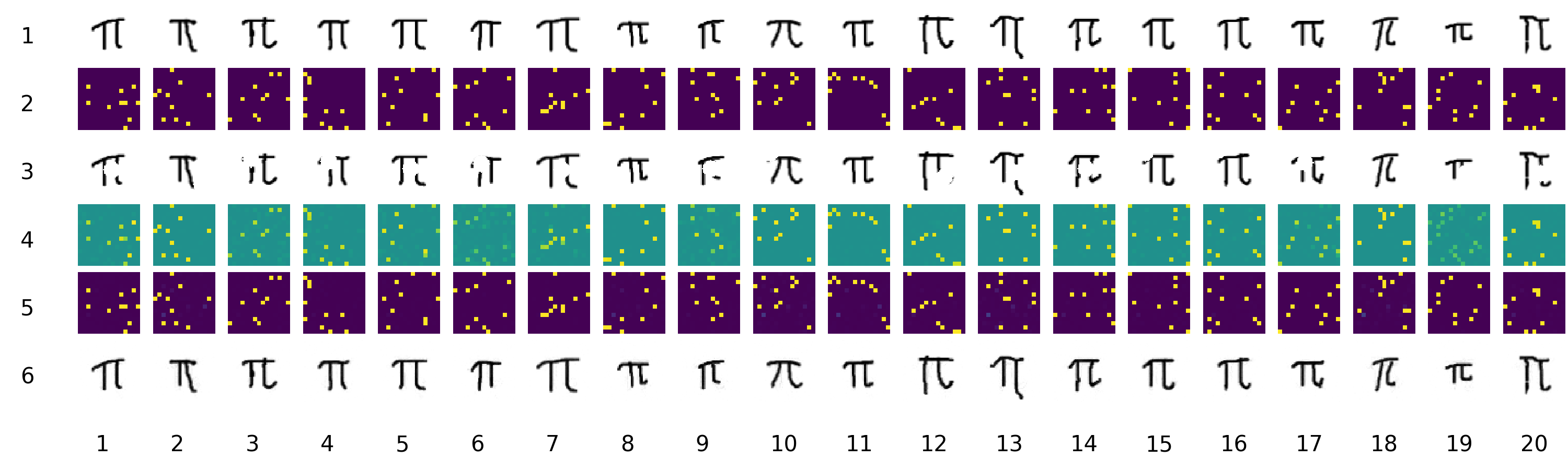}
		\caption{With occlusion (fraction=0.3).}\label{fig:instance_images_occl_03}
	\end{subfigure}
	\bigskip
	\begin{subfigure}{\textwidth}
  		\centering
    	\includegraphics[width=\textwidth]{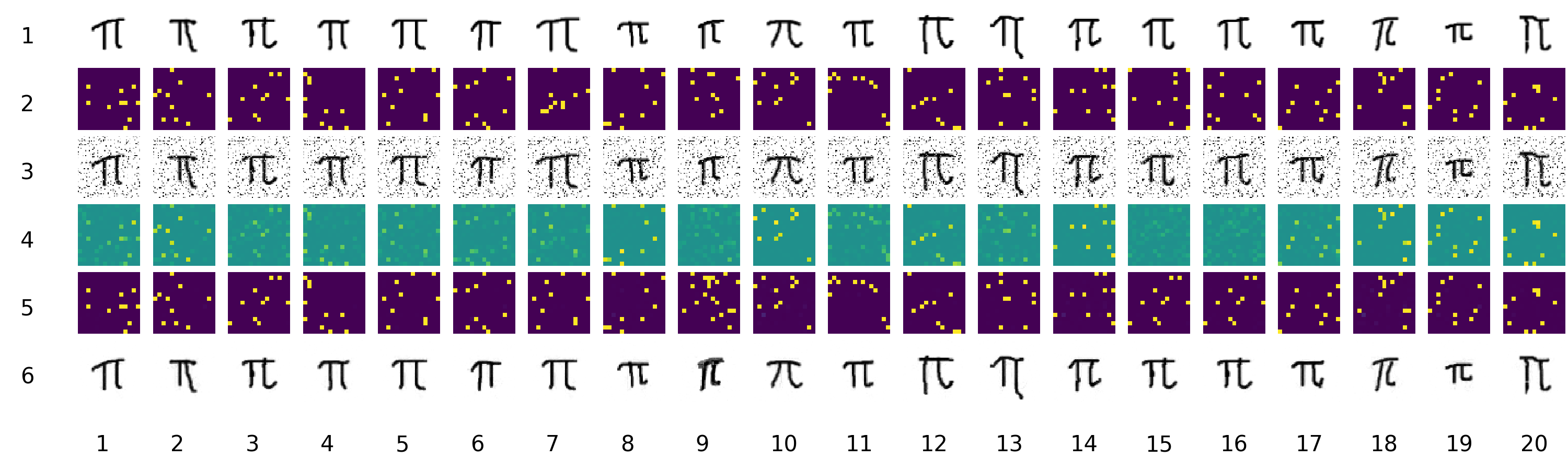}
		  \caption{With noise (fraction=0.3).}\label{fig:instance_images_noise_03}
	\end{subfigure}
  	\caption{\textbf{\OneshotInst\ test patterns as they propagate through AHA}. See Table~\ref{table:legend} for legend.}
  	\label{fig:instance_qual}
\end{figure*}

\begin{figure*}[ht]
	\begin{subfigure}{\textwidth}
		\centering
		\includegraphics[width=\textwidth]{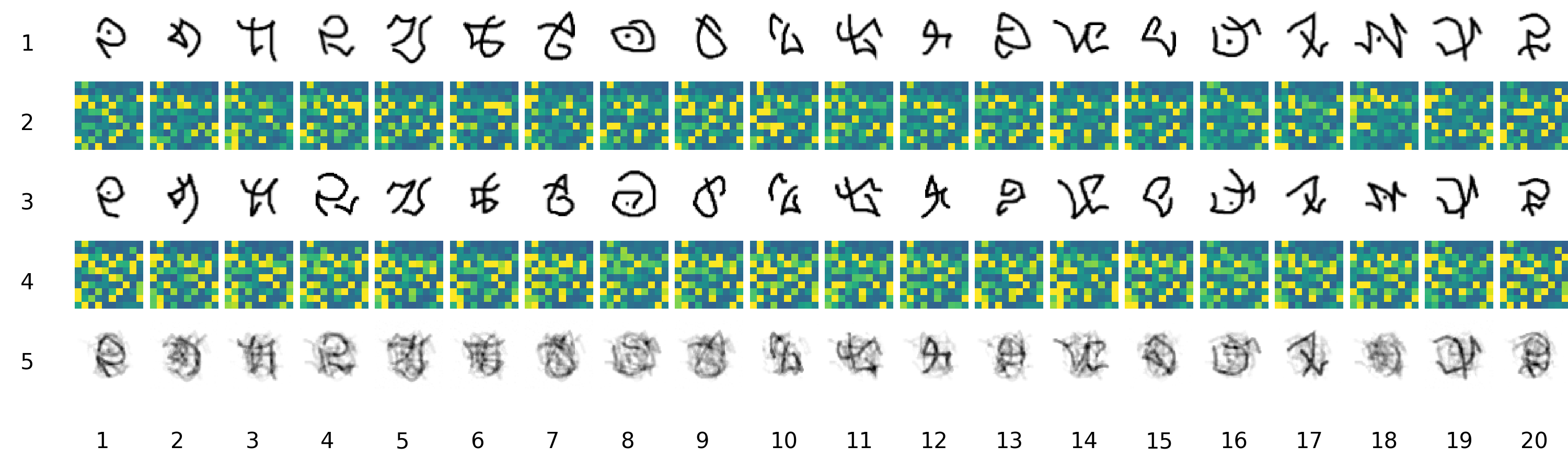}
	  \caption{No occlusion or noise.}\label{fig:fae_oneshot}
	\end{subfigure}
	\bigskip
	\begin{subfigure}{\textwidth}
		\centering
    	\includegraphics[width=\textwidth]{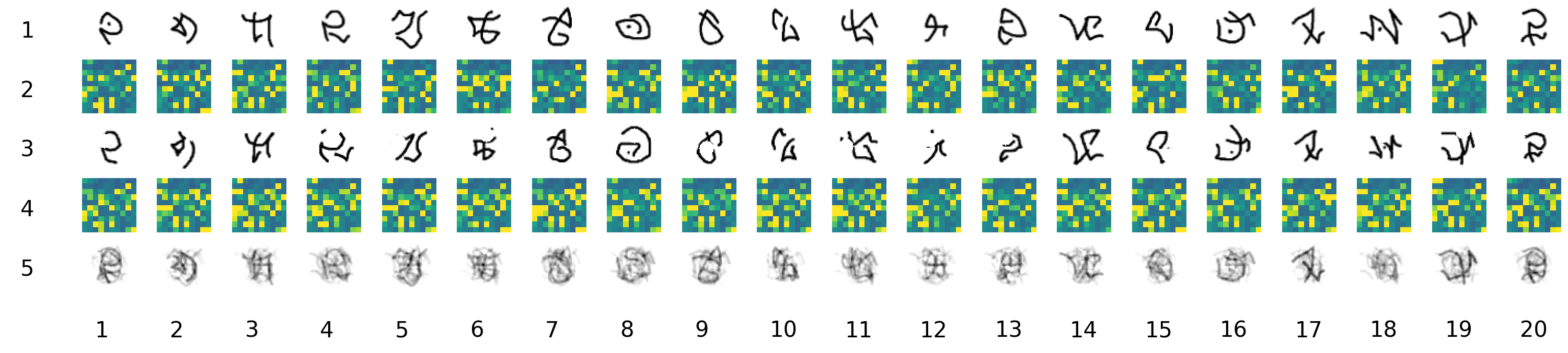}
		\caption{With occlusion (fraction=0.3).}\label{fig:fae_oneshot_occl}
	\end{subfigure}
	\bigskip
	\begin{subfigure}{\textwidth}
  		\centering
    	\includegraphics[width=\textwidth]{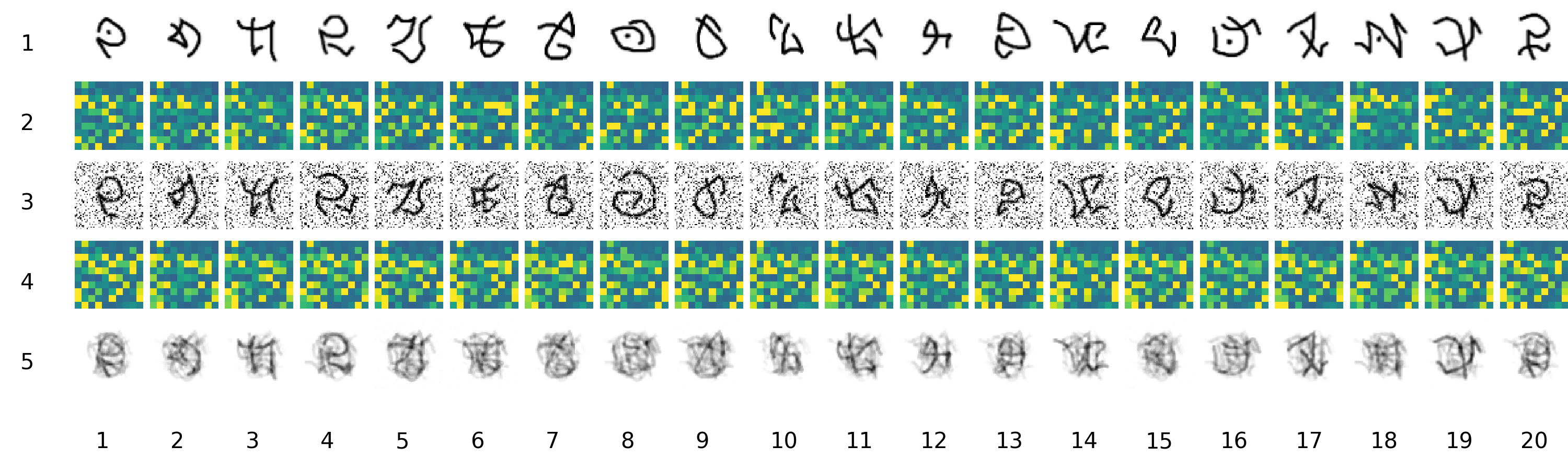}
		  \caption{With noise (fraction=0.3).}\label{fig:fae_oneshot_noise}
	\end{subfigure}
	\bigskip
  	\caption{\textbf{\OneshotGen\ test patterns as they propagate through \FNN.} Row 1: Train images, Row 2: Test images, Row 3: PM output, the image reconstruction.}
  	\label{fig:fae_oneshot_all}
\end{figure*}

\begin{figure*}[ht]
	\begin{subfigure}{\textwidth}
		\centering
		\includegraphics[width=\textwidth]{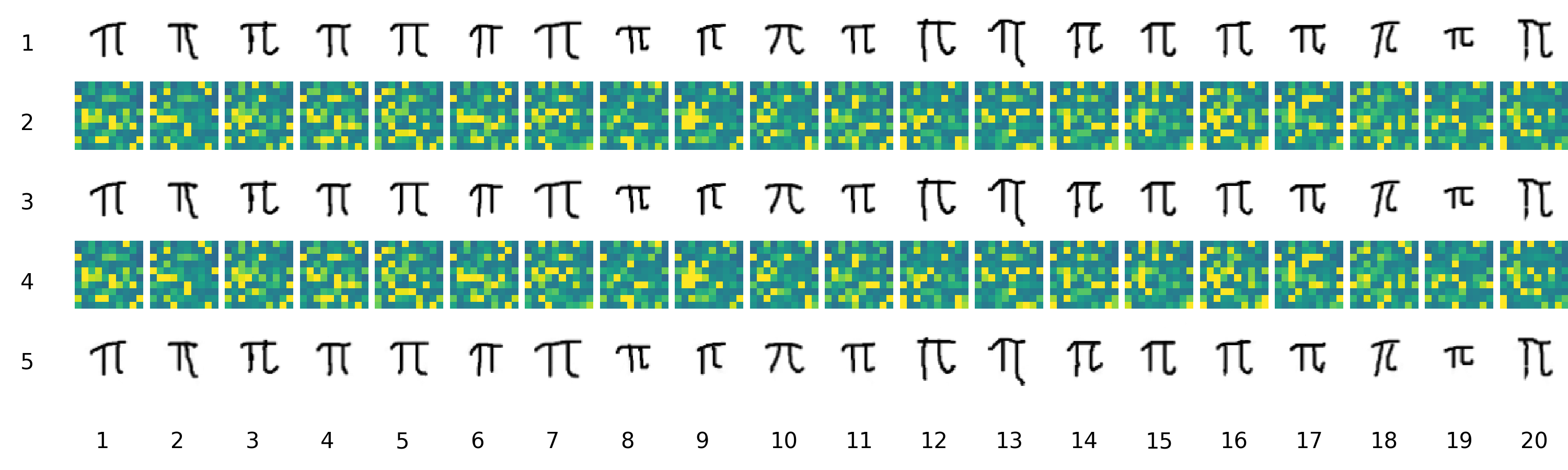}
	  \caption{No occlusion or noise.}\label{fig:fae_instance}
	\end{subfigure}
	\bigskip
	\begin{subfigure}{\textwidth}
		\centering
    	\includegraphics[width=\textwidth]{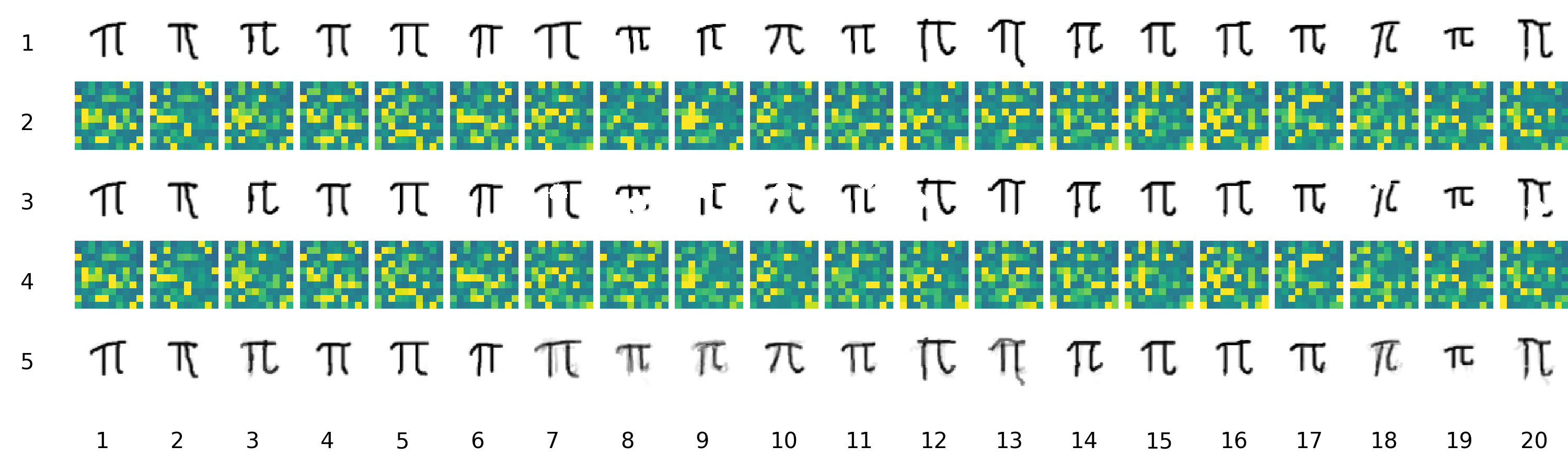}
		\caption{With occlusion (fraction=0.3).}\label{fig:fae_instance_occl}
	\end{subfigure}
	\bigskip
	\begin{subfigure}{\textwidth}
  		\centering
    	\includegraphics[width=\textwidth]{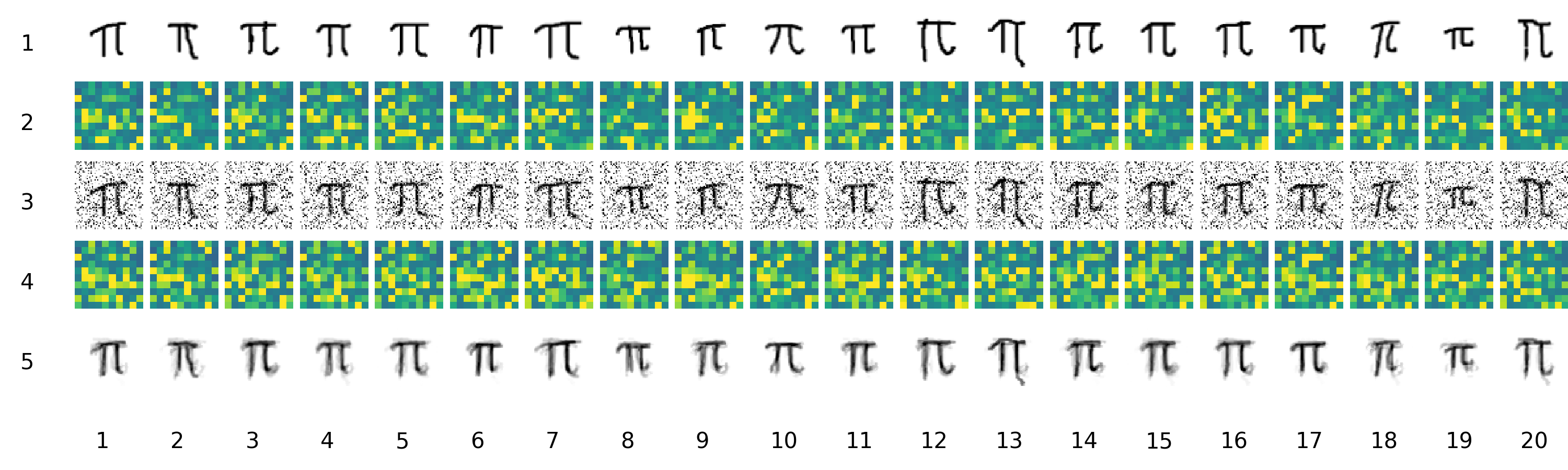}
		  \caption{With noise (fraction=0.3).}\label{fig:fae_instance_noise}
	\end{subfigure}
	\bigskip
  	\caption{\textbf{\OneshotInst\ test patterns as they propagate through \FNN.} Row 1 shows Train images, Row 2 Test images, and Row 3 shows the PM output, the VC reconstruction.}
  	\label{fig:fae_instance_all}
\end{figure*}

Internal signals are shown as they propagate through the components of STM implementations for typical scenarios. For AHA they are shown in Figures \ref{fig:class_qual} and \ref{fig:instance_qual}. A legend is given with Table~\ref{table:legend}. 
The same is shown for \FNN\ in Figures \ref{fig:fae_oneshot_all} and \ref{fig:fae_instance_all}. 
Additional figures showing high and extreme levels of image corruption are included in Appendix~\ref{app:high_corruption}.

\begin{table}[ht]
\begin{center}
\begin{tabular}{ l l }
\hline
\textbf{Row} & \textbf{Pattern} \\ \hline
1 & Train samples \\ \hline
2 & \PS\ output \\ \hline
3 & Test samples \\ \hline
4 & \PR\ output \\ \hline
5 & \PC\ output \\ \hline
6 & \PM\ output (image reconstruction) \\ \hline
\end{tabular}
\caption[Table caption text]{Legend for Figures \ref{fig:generalization_images} to \ref{fig:instance_images_noise_03}.}
\label{table:legend}
\end{center}
\end{table}

For LTM+AHA, \PS\ (row 2) produces non-overlapping patterns from the train batch samples, even where the samples are extremely similar (different exemplars of the same class). This shows that pattern separation is working as intended. 
For test batch samples, \PR\ outputs are in most cases, recognisable retrievals of the correct \PS\ pattern. However they are noisy and unevenly attenuated. Given this input, \PC\ then converges to a crisp recall of one of the memorised patterns, a basin of attraction in the Hopfield network, but not always the correct one. There was one rare observed exception where \PC\ produced a superposition of states.
\PM\ produced sharp complete versions using the \PC\ pattern, resulting in complete recalled patterns.

In the presence of occlusion, complete train samples are still recalled (even if they are not the correct sample).
In addition, the retrieval is mainly correct even when the occluded portion disrupts the topology of the sample. For example, in the case of occlusion, Figure~\ref{fig:generalization_images_occl}, columns 2, 5, 6, 7, 11, 12, 14, 16, 19 and 20 compared to error in column 8. 
In the case of noise, Figure~\ref{fig:instance_images_occl_03}, see examples in columns 1, 3, 4, 5, 6, 7, 9, 12, 13, 14, 17, 19 and 20 with no errors.

Recall tends to fail gracefully, recalling similar samples. For example, column 8 Figure~\ref{fig:generalization_images_occl}. The recalled character has a large circular form with a dot in the middle. It shares many of the feature cues with the correct character.
In other failure cases such as column 13 and 15, Figure~\ref{fig:generalization_images_occl}, occlusion has not damaged the sample, but \PR\ appears to have recalled multiple patterns that result in a superposition or converge to the wrong train sample respectively. 
It is not clear from visual inspection of the train and test samples, if ambiguous visual features have contributed to the error, or if it is due to the representational and generalisational capacity of \PR.  

For LTM+\FNN, the recalled images are blurred with substantial ghosting from other character forms for \oneshotGen. For \oneshotInst, the same phenomenon is observed, but to a far lesser extent and the characters are relatively clear.

\section{Discussion}
This section begins with a general discussion of the results, followed by sub-sections on particular themes.

The results demonstrate that AHA performs replay with separation and completion in accordance with the expected behaviour of the hippocampus \citep{OReilly2014, Rolls2018, Kandel1991}.
AHA performs well at distinguishing similar inputs, and unlike previous CLS hippocampal models, also performs well given large observational variation, and data that originates from grounded sensory input.
The variation can result from seeing different exemplars of the learnt class (e.g. I see a fork for the first time and now I can recognise other forks) or from corruption (e.g. looking for the same fork, but it is dirty or partially occluded). 
The former condition is more difficult to test without the use of grounded sensor input and modelling neurons at a higher level of abstraction.
Emphasis in AHA on the EC-CA3 network and interplay with DG-CA3 is likely to play a significant role in this capability (see section~\ref{sec:theory_of_operation_collab} below).

The use of a complementary STM (LTM+STM) outperforms the baseline (LTM) across all performance measures and tasks. Our results validate that this complementary architecture can perform a wider range of hippocampal functions.
The AHA-STM model outperforms the \FNN-STM.
The advantage in accuracy is most significant for generalisation and in the presence of occlusion.
AHA-STM recall outputs are cleaner and more specific even if they are incorrect, up to very high levels of image corruption, especially where generalisation is required.
We hypothesise that this type of recall is significantly more useful for the purpose of consolidating few-shot exposures into a complementary LTM, analogous to the biological complementary learning systems.
The advantages of AHA over \FNN\ are justification for the use of a more complex heterogeneous architecture.

Given that \PR\ makes a crucial contribution to AHA's accuracy (see section~\ref{sec:import_pr} below), perhaps it is not surprising that \FNN, which is also a 2-layer FC-ANN trained in a similar manner (albeit different size and configuration), can do almost as well.
However, the interaction of pathways in AHA provides an accuracy boost, and most importantly allows the crisp, \textit{opinionated} recall.

LTM+AHA is competitive to conventional ANN techniques on the standard Omniglot Benchmark conditions (which measures accuracy on the \oneshotGen\ task without corruption).
Referring to Table~\ref{table:comparison}, BPL and RCN are significantly ahead of other methods, and are similar to human performance. This is expected for BPL, as it exploits domain-specific prior knowledge about handwriting via stroke formation. RCN, by virtue of the design which is modelled on the visual cortex, is also specialised for this type of visual task. It is less clear how it could be applied to other datasets and problems where contours are less distinct or less relevant.
The Simple Conv Net (CNN) represents a standard approach for deep learning. 
AHA is equally good despite using biologically plausible local credit assignment, fewer computational resources, and no labels. Additionally, AHA demonstrates a broader range of capabilities such as pattern separation and replay.

Overall, the results demonstrate that AHA has potential to augment conventional ML models to provide an `episodic' \oneshot\ learning capability as tested in this study, but even more exciting is the prospect of using the replay capability to integrate that knowledge into the model as long-term knowledge.

\subsection{Advantage of Using a Complementary STM}
Due to the role that LTM plays, providing primitive concepts, it is only able to complete primitives rather than the `episode' or conjunction of primitives. In addition it does not have a memory for multiple images, so there is no way for it to recognise a specific example, limiting accuracy and making it unable to recall. 

\subsection{Importance of \PR\ and \PC}
\label{sec:import_pr}

The accuracy results demonstrate that \PR\ performs classification significantly better than \PC.
This is somewhat surprising, given the conventional view that CA3 performs the bulk of pattern completion \citep{McClelland1995, Norman2003, Ketz2013, Schapiro2017a}.
As \PR\ learns to associate overlapping representations (in \VC) to complete target patterns (from \PS), 
\PR\ would produce complete patterns as outputs.
Visual inspection shows that it does produce largely complete but noisy patterns.
It is further evidence that \PR\ is well suited to be primarily responsible for recall from CA3 \citep{Rolls2013}.
\PC\ does fulfil a vital role for additional completion and sharpening so that the pattern can be effectively reconstructed by \PM\ for reinstatement of the original cortical representation. 

It is possible that previous computational studies \citep{Norman2003, Ketz2013, Greene2013, Schapiro2017a} did not encounter this discrepancy because experiments focused on a narrower problem set of pattern separation, in which AHA-\PC performed much closer to \PR.
There are other factors to consider such as the division of EC into discrete receptive fields and capacity difference between \PR\ and \PC\ equivalent networks \citep{Rolls2013}. 
However, the importance of \PR\ in AHA suggests that the connectivity of EC-CA3 is more important than previously acknowledged.

Note that a small and less significant source of accuracy bias toward \PR\ occurs in the cases where \PR\ outputs a superposition of possible patterns, enhancing the chance of a correct match via MSE. In contrast, \PC\ is designed to retrieve a single, sharp complete sample and in doing so is unable to hedge its bets.

\subsection{Learning is Task Independent}
The episodic representations in AHA can be used for different tasks.
For the \oneshotInst\ test, the symbol generalises over versions of the same exemplar (subject to noise and occlusion). This is the standard definition of episodic learning in previous studies \citep{Norman2003, Ketz2013, Greene2013, Schapiro2017a, Rolls2013}.
For the \oneshotGen\ test, the symbol learnt generalises over multiple exemplars of the same class (additionally subject to noise and occlusion). This is the standard definition for generalisation in classification. 
In both cases, the symbol represents a conjunction of visual primitives from the same set, and competence at both tasks is accomplished by unification of separation and completion.

In reality the boundary between what is a class and what is the exemplar is continuous, subjective and in many circumstances depends on the task. 
For example, you could define the character itself as a class, and the corrupted samples are exemplars. Another example is a Labrador dog, the class could be the animal type - dog (another exemplar would be cat), or the breed - Labrador (another exemplar would be Poodle). 
AHA demonstrates this flexibility to the task by accomplishing both \oneshotGen\ and \oneshotInst.
It is a characteristic we can expect from the hippocampus through observation of everyday animal behaviour.

\subsection{Learning is Hierarchical}
According to our principle of operation, the flexibility explained above should extend to any level of abstraction i.e. learning the particular details of a spoken sentence to the simple fact that you had a conversation. AHA learns an episode, a conjunction of primitives, and then generalises over variations in that combination. The meaning of the episode depends on the level of abstraction of the primitives. 
This ties into the discussion about iteratively building more abstract concepts in Theory of Operation (Section~\ref{sec:theory_of_operation}).
One possibility to exert control over the level of detail encoded would be an attentional mechanism, mediating the selection of primitives.

\section{Conclusions}
This paper presented AHA, a novel computational model of the hippocampal region and subfields (functional and anatomical sub-regions). AHA uses biologically-plausible, local learning rules without external labels.
AHA performs fast learning, separation and completion functions within a unified representation. The symbolic representations generated within AHA can be grounded - mapped back to the original input.
We describe how this architecture complements the incremental statistical learning of popular ML methods. AHA could extend their abilities to more animal-like learning, such as \oneshot\ learning of a conjunction of primitives (an episode). This could enable ML to perform more sample-efficient learning, and reason about specific instances.

The system was tested on visual recognition tasks featuring grounded sensor data.
We posed a new benchmark based on the visual \oneshot\ classification task in \citet{Lake2015}.
An additional \oneshotInst\ test was introduced, testing the ability to reason about specific \instance/s. We also added image corruption with noise and occlusion, and all experiments were repeated several times to evaluate consistency.
The results show that the subfields' functionality matches biological observations, and demonstrates a range of capabilities.
AHA can memorise a set of samples, learn in \oneshot, perform classification requiring generalisation and identify specific instances (reason about specifics) in the face of occlusion or noise. It can accurately reconstruct the original input samples which can be used for consolidation of long-term memories to influence future perception. 
AHA \oneshot\ classification accuracy is comparable to existing ML methods that do not exploit domain-specific knowledge about handwriting.

The experiments expanded the scope of previous biological computational model studies, shedding light on the role and interplay of the subfields and aiding in understanding functionality of the hippocampal region as a whole.

\section{Future Work}
In future work, we will explore two ways that AHA could augment incremental learning ML models. Firstly, the use of grounded non-symbolic \VC\ reconstructions from \PM\ to selectively consolidate memories so that they can affect future perception. 
Secondly, AHA could directly and immediately augment slow-learning ML models by interpolating rapidly learned classification or predictions from AHA with the incremental learning model. This approach is compatible with a wide variety of models and would make them more responsive to rapidly changing data, or where fewer labelled samples are available.
We would also like to investigate how these representations can be fed back through AHA in `big-loop' recurrence to learn statistical regularities across episodes (see Section~\ref{sec:comp_model}) and to resolve ambiguous inputs.

\subsection*{Acknowledgments}
A big thank-you to Elkhonon Goldberg for enriching discussions on the hippocampal region, its relationship to the neocortex and their role in memory.
We also greatly appreciate the insight that Rotem Aharon provided in analysing and improving the dynamics of Hopfield networks.
Matplotlib \citep{Hunter2007a} was used to generate Figures~\ref{fig:generalization_images} to \ref{fig:instance_images_noise_03} and Figure~\ref{fig:vc_filters}.

\subsection*{Author Contributions}
GK and DR devised the concept and experiments. GK, DR and AA wrote the code. GK, DR and AA executed the experiments. GK, DR and AA wrote the paper.

\bibliography{bibliography.bib}
\bibliographystyle{apa}

\appendix

\section{\PC\ Input Signal Conditioning}
\label{sec:app_pc}

\subsection{\PS-\PC\ Memorisation Input}
\label{sec:app_ps_pc}
The \PS\ output $X'$ is conditioned for memorisation into the Hopfield network, which benefits from binary vectors in the range $[-1, 1]$. The conditioning function is:
	\begin{gather*}
		X'=2\sign(X)-1
	\end{gather*}
It is implied that $X$ must be unit range.

\subsection{\PR-\PC\ Retrieval Input}
\label{sec:app_pr_pc}
The \PR\ output $Y'$ is optimised for the classification task. To present a valid cue for Hopfield convergence, additional conditioning is required. 
First, a norm is applied per sample in a batch of size $K$, with a gain term $\gamma=10$.
	\begin{gather*}
		Z_{i} = \gamma \cdot Y_{i} \cdot (1 / \sum_{j=1}^{K} Y_{j})
	\end{gather*}
Next, the range is transformed to the Hopfield operating range of $[-1, 1]$ and finally an offset $\Theta$ is applied per sample. Intuitively, $\Theta$ shifts the output distribution to straddle zero, with at least $k$ bits $> 0$, where $k$ is the fixed sparsity of the stored patterns from \PS.
	\begin{gather*}
		Y'=(2Z-1)+ \Theta
	\end{gather*}
Given that it is in range $[-1, 1]$, anything negative acts as inhibition, so the balance is very important. The inputs are relatively sparse, dominated by background (negative). To allow some elements to reach an active status before the many inhibitory elements dominate, it's necessary to initialise the distribution as described.

\section{\VC}
\label{sec:app_vc}

\subsection{Sparse convolutional autoencoder}
\label{sec:app_scae}
Our sparse convolutional autoencoder is based on the winner-take-all autoencoder \citep{Makhzani2015}. 
To select the top-$k$ active cells per mini-batch sample, we use a convolutional version of the original rule from \citep{Makhzani2013}.
The top-$k$ cells are selected independently at each convolutional position by competition over all the filters.
In addition, a lifetime sparsity rule is used to ensure that each filter is trained at least once per mini-batch (i.e.
a lifetime of $1$ sample per mini-batch).
We found that a single autoencoder layer with tied weights was sufficient for the Omniglot character encoding. However, additional layers could have been trained with local losses without violating our biological plausibility rules. To reduce the dimensionality of the \VC\ output, we applied max-pooling to the convolutional output.

\subsection{Pre-training}
\begin{figure}[ht]  
  \centering
    \includegraphics[width=0.48\textwidth]{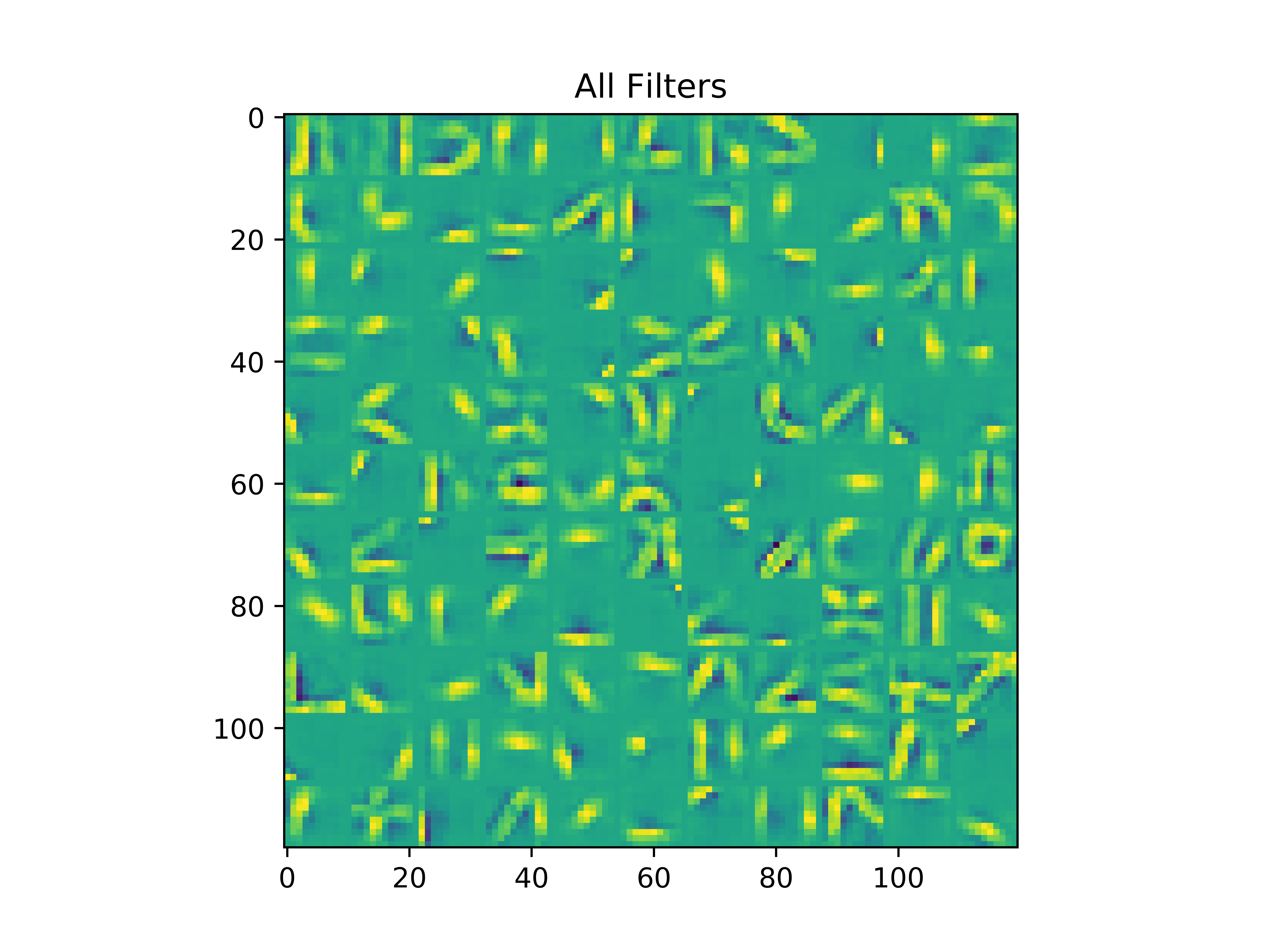}
  \caption{\VC\ sparse convolutional autoencoder filters.}\label{fig:vc_filters}
\end{figure}

Pre-training of the sparse convolutional autoencoder develops filters that detect a set of primitive visual concepts that consist of straight and curved edges, sometimes with junctions (Figure~\ref{fig:vc_filters}).

\subsection{Interest Filter}
\label{sec:app_vc_if}
\begin{figure}
  \centering
    \includegraphics[width=0.98\columnwidth]{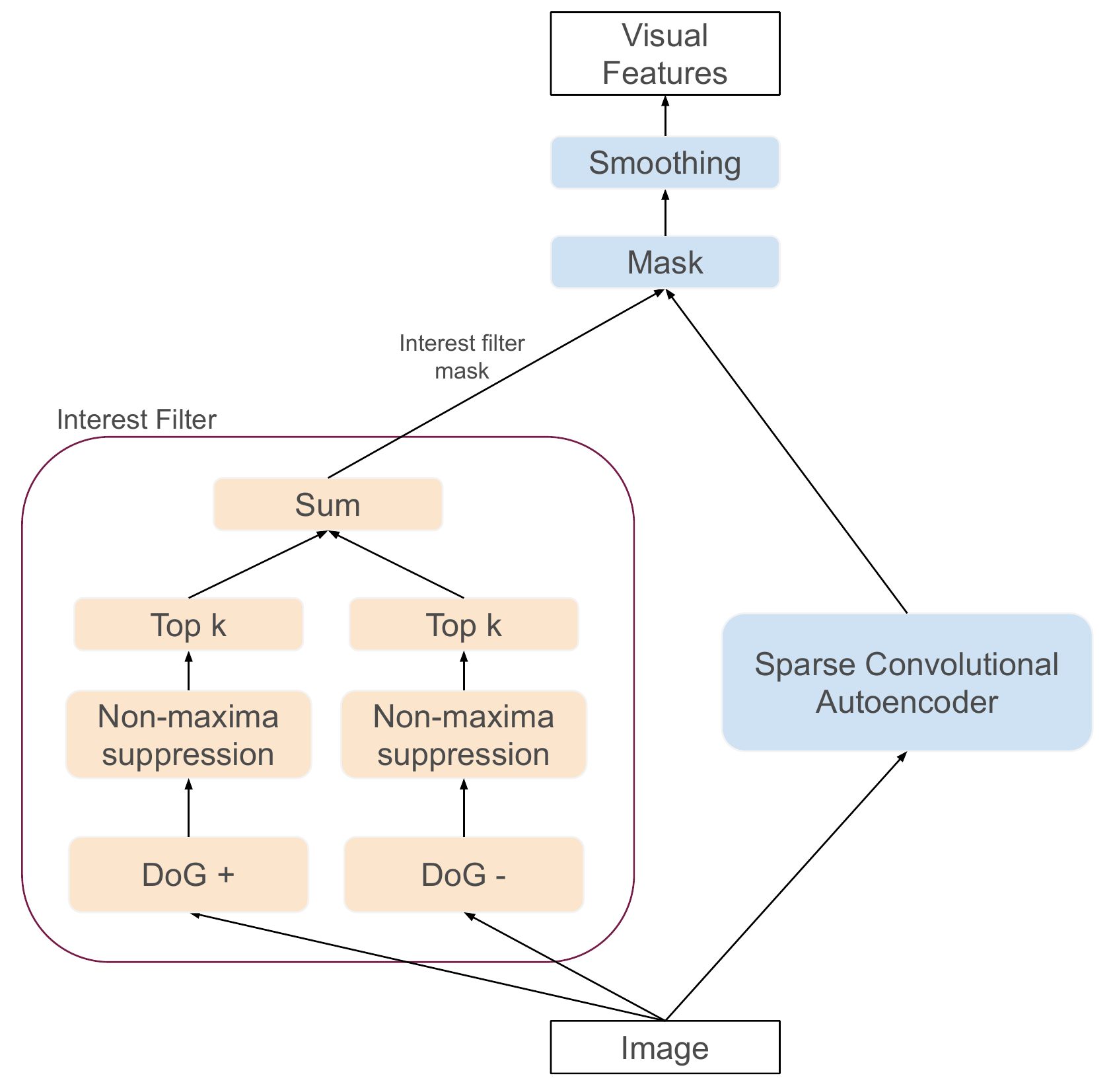}
  \caption{Architecture of the VC including `Interest Filter'. The VC is a single layer sparse convolutional autoencoder with masking to reduce the magnitude of background features.}\label{fig:vc}
\end{figure}

As shown in Figure~\ref{fig:vc}, positive and negative DoG filters are used to enhance positive and negative intensity transitions. The filter output is subject to local non-maxima suppression to merge nearby features and a `top-$k$' function creates a mask of the most significant features globally.
Positive and negative masks are combined by summation giving a final `Interest Filter' mask that is applied to all channels of the convolutional output volume.
A smoothing operation is then applied to provide some tolerance to feature location.
There is a final max-pooling stage to reduce dimensionality.
The non-maxima suppression and smoothing are achieved by convolving Gaussian kernels with the input. Parameters are given in Table~\ref{table:params}.

\section{System Design}
\label{app:app_system}

\begin{figure*}
  \centering
    \includegraphics[width=0.95\textwidth]{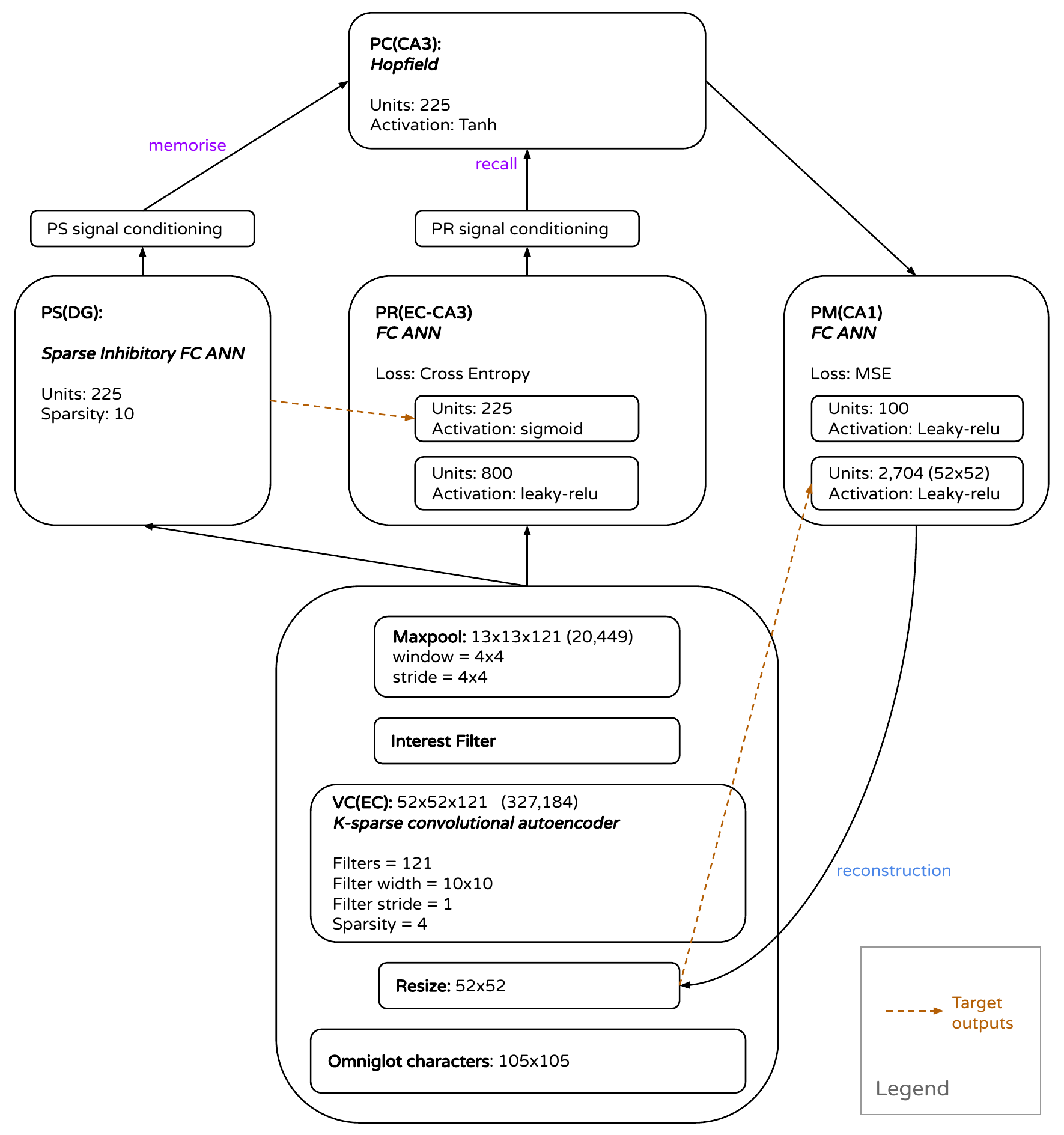}
  \caption{Architecture details (the hyperparameters shown are for the experiments after pre-training the VC).}\label{fig:arch_details}
\end{figure*}

The full architecture of AHA is shown in Figure~\ref{fig:arch_details}. The hyperparameters used in our experiments are shown in Table~\ref{table:params}. We used the Adam optimizer \citep{Kingma2013} in all experiments.

\begin{table*}
\begin{center}
\begin{tabular}{ | l | c | }
\hline
\textbf{\PS\ - Pattern Separation with Inhibitory Sparse Autoencoder} &  \\ \hline
$k$ (sparsity) & 10 \\ \hline
$h$ (number of units) & 225 \\ \hline
$\gamma$ (inhibition decay rate) & 0.95 \\ \hline
$\upsilon$ (weight knockout rate) & 0.25 \\ \hline
\textbf{\PC\ - Pattern Completion with Hopfield Network} &  \\ \hline
$\lambda$ (gain) & 2.7 \\ \hline
$n$ (cells updated per step) & 20 \\ \hline
$N$ (iterations) & 70 \\ \hline
$h$ (number of units) & 225 \\ \hline
\textbf{\PR\ - Pattern Retrieval with Fully-connected ANN} &  \\ \hline
$\eta$ (learning rate) & 0.01 \\ \hline
$h$ (number of hidden units) & 1000 \\ \hline
$o$ (number of output units) & 225 \\ \hline
$\lambda$ (L2 regularisation) & 0.000025 \\ \hline
\textbf{\PR\ - Signal Conditioning} &  \\ \hline
$\gamma$ (gain) & 10 \\ \hline
\textbf{\PM\ - Pattern Mapping with Fully-connected ANN} &  \\ \hline
$\eta$ (learning rate) & 0.01 \\ \hline
$h$ (number of hidden units) & 100 \\ \hline
$o$ (number of output units) & 100 \\ \hline
$\lambda$ (L2 regularisation) & 0.0004 \\ \hline
\textbf{\VC\ - Vision Component Sparse Convolutional Autoencoder} & \emph{()=pre-training} \\ \hline
$\eta$ (learning rate) & 0.001 \\ \hline
$k$ (sparsity) & (1), 4 \\ \hline
$f$ (number of filters) & 121 \\ \hline
$f_{w}$ (filter width) & 10 \\ \hline
$f_{h}$ (filter height) & 10 \\ \hline
$f_{s}$ (filter stride) & (5), 1 \\ \hline
Batches (pre-training) & 2000 \\ \hline
Batch size (pre-training) & 128 \\ \hline
\textbf{\VC\ - Vision Component Interest Filter} &  \\ \hline
DoG kernel size & 7 \\ \hline
DoG kernel std & 0.82 \\ \hline
DoG kernel k & 1.6 \\ \hline
Non-maxima suppression size & 5 \\ \hline
Non-maxima suppression stride & 1 \\ \hline
Smoothing size & 15 \\ \hline
Smoothing encoding std & 2.375 \\ \hline
k (number of features) & 20 \\ \hline
\textbf{\VC\ - Vision Component} &  \\ \hline
Resize & 0.5 \\ \hline
Max-pooling size & 4  \\ \hline
Max-pooling stride & 4  \\ \hline
\textbf{\FNN\ - Baseline STM} & \\ \hline
$\eta$ (learning rate) & 0.01 \\ \hline
$h$ (number of hidden units) & 100 \\ \hline
$\lambda$ (L2 regularisation) & 0.00004 \\ \hline
\end{tabular}
\caption[Table caption text]{Hyperparameter values for reported experiments.}
\label{table:params}
\end{center}
\end{table*}
 
\section{High and Extreme Image Corruption}
\label{app:high_corruption}

\begin{figure*}[ht]
     \centering
     \begin{subfigure}{\textwidth}
         \centering
	 	 \includegraphics[width=\textwidth]{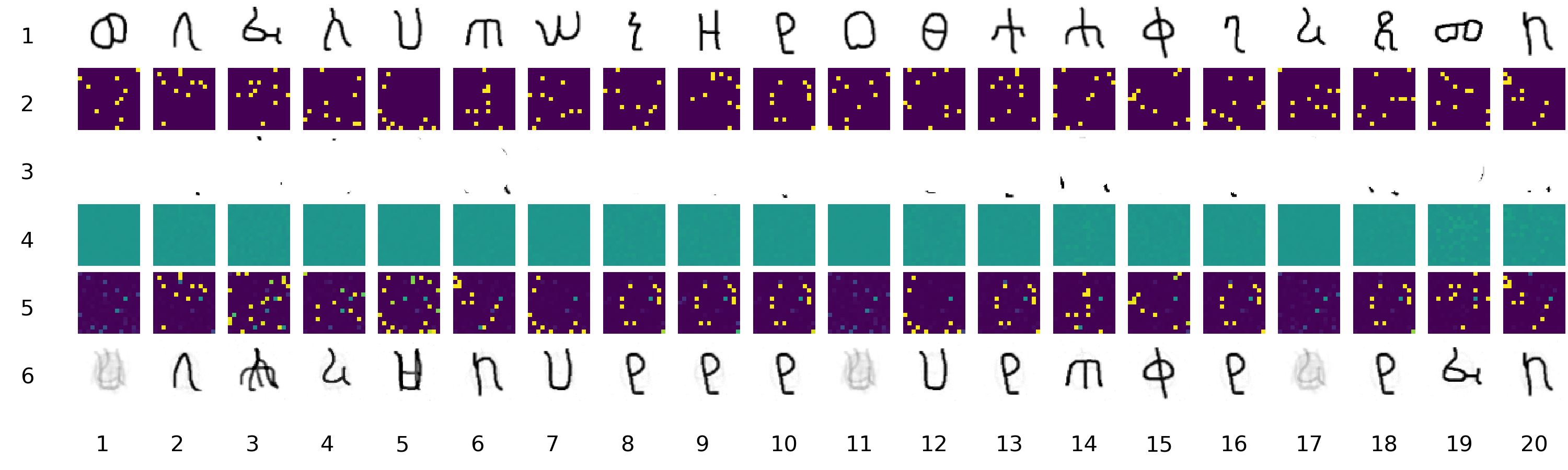}
		 \caption{\OneshotGen\ and Occlusion}
     \end{subfigure}
     \hfill
     \begin{subfigure}{\textwidth}
         \centering
	 	 \includegraphics[width=\textwidth]{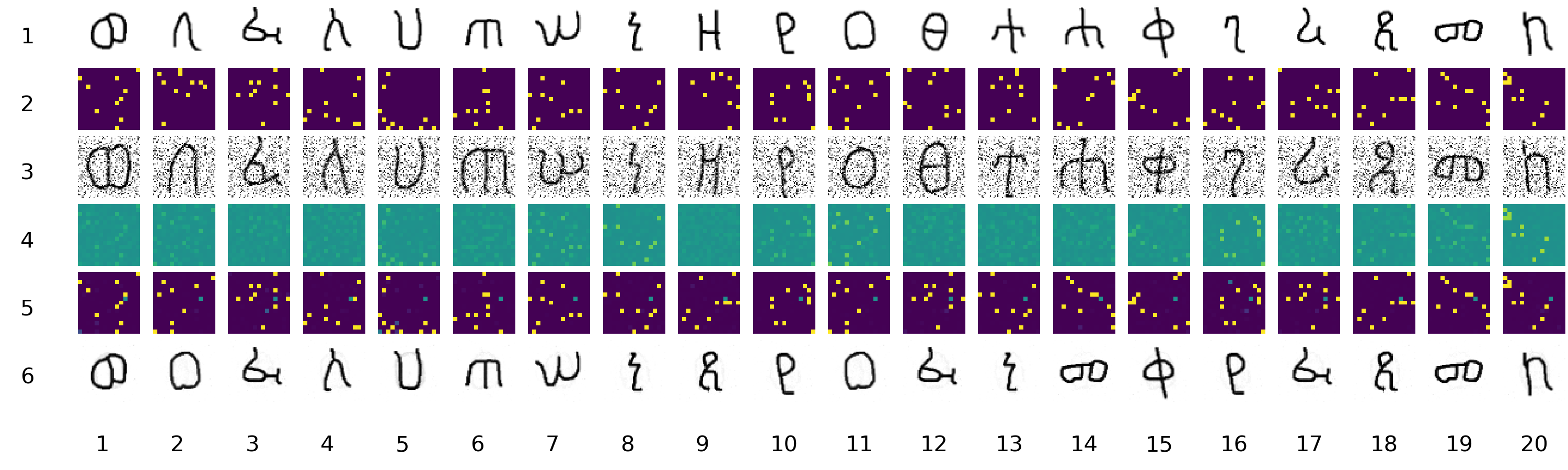}
  		 \caption{\OneshotGen\ with Noise}
     \end{subfigure}
     \begin{subfigure}{\textwidth}
         \centering
	 	 \includegraphics[width=\textwidth]{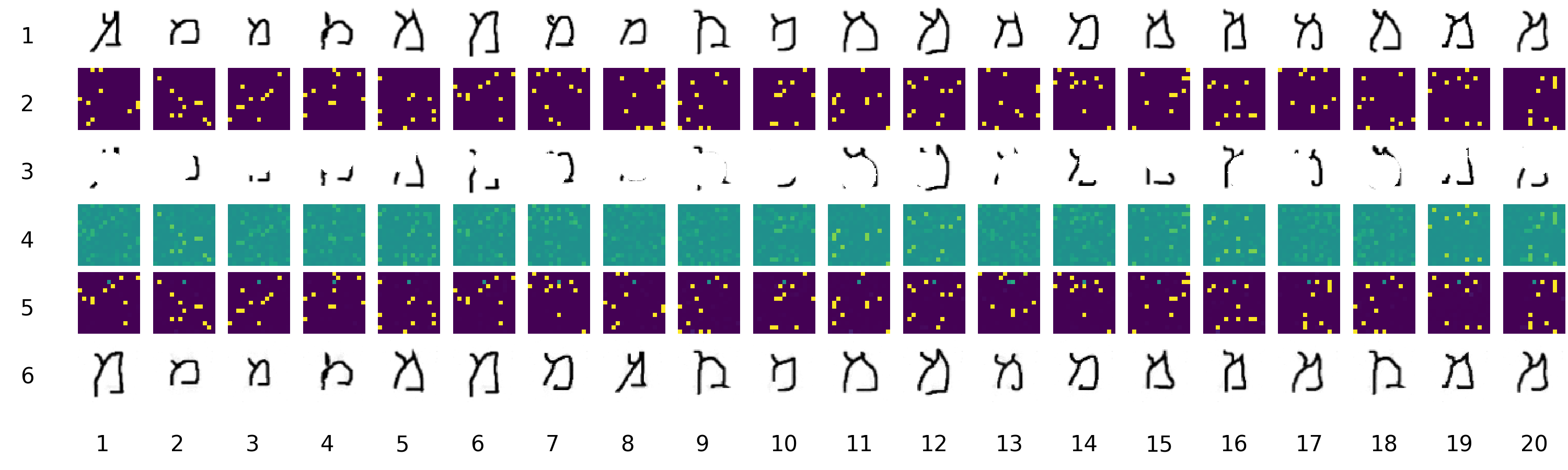}
		 \caption{\OneshotInst\ with Occlusion}
     \end{subfigure}
     \hfill
     \begin{subfigure}{\textwidth}
         \centering
	 	 \includegraphics[width=\textwidth]{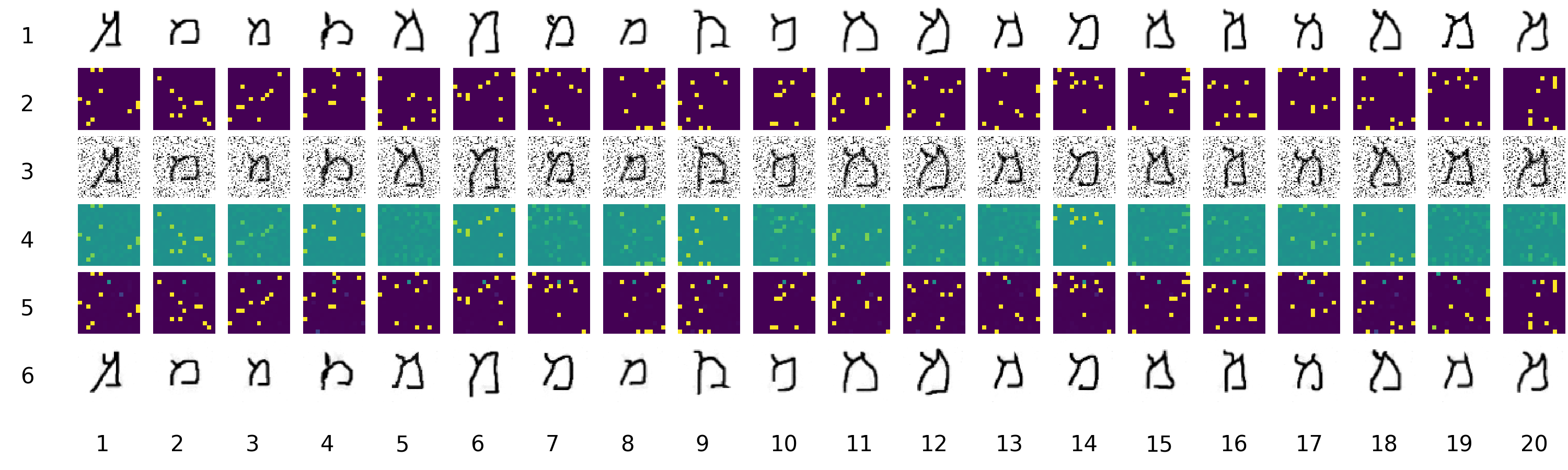}
  		 \caption{\OneshotInst\ with Noise}
     \end{subfigure}
     \caption{AHA at high level of image corruption ($=0.6$).}
     \label{fig:AHA_high_corruption_06}
\end{figure*}

\begin{figure*}[ht]
     \centering
     \begin{subfigure}{\textwidth}
         \centering
	 	 \includegraphics[width=\textwidth]{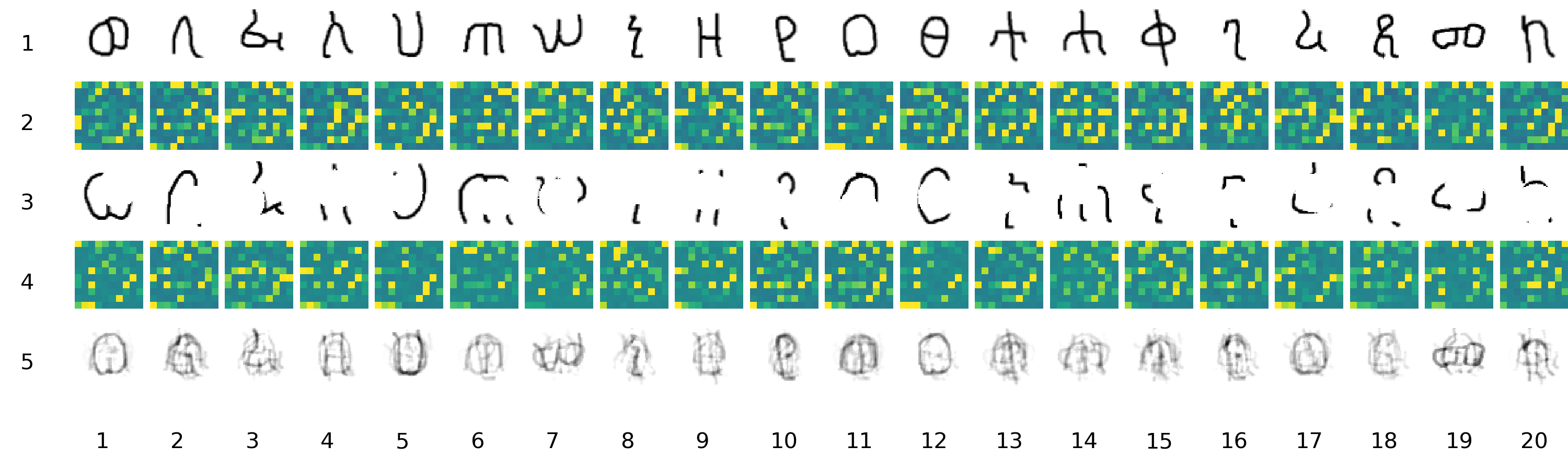}
		 \caption{\OneshotGen\ with Occlusion}
     \end{subfigure}
     \hfill
     \begin{subfigure}{\textwidth}
         \centering
	 	 \includegraphics[width=\textwidth]{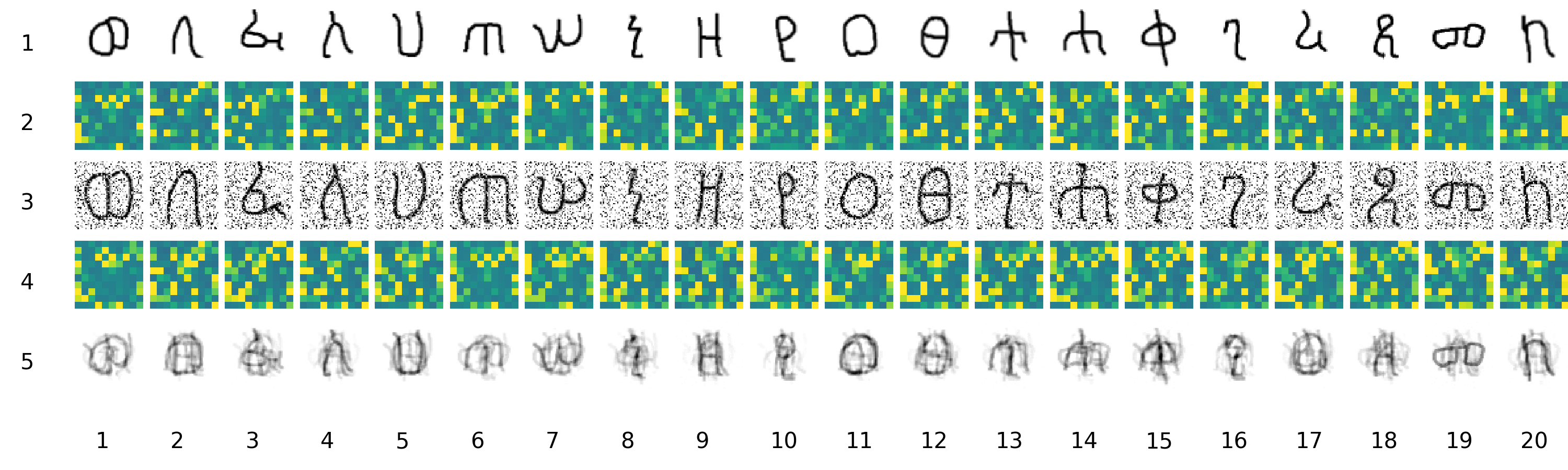}
  		 \caption{\OneshotGen\ with Noise}
     \end{subfigure}
     \begin{subfigure}{\textwidth}
         \centering
	 	 \includegraphics[width=\textwidth]{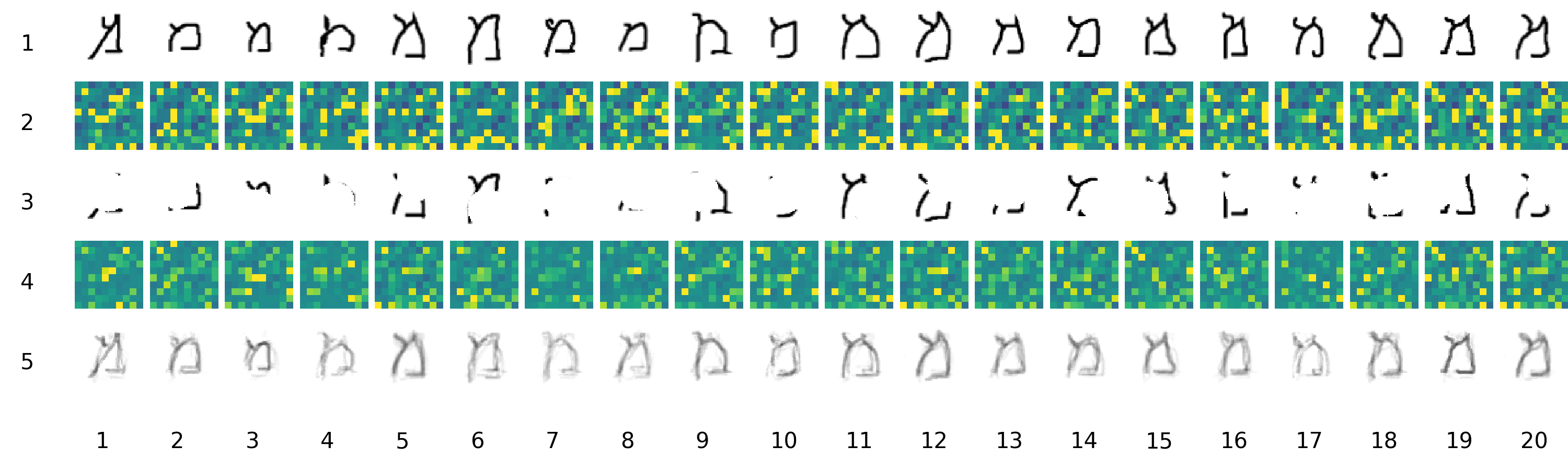}
		 \caption{\OneshotInst\ with Occlusion}
     \end{subfigure}
     \hfill
     \begin{subfigure}{\textwidth}
         \centering
	 	 \includegraphics[width=\textwidth]{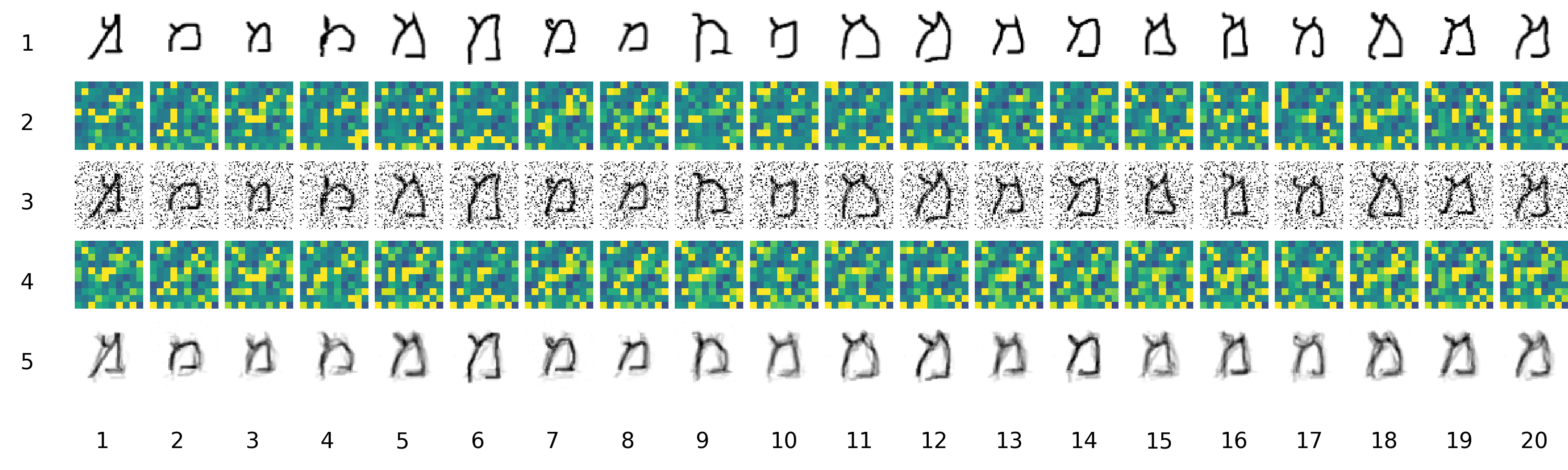}
  		 \caption{\OneshotInst\ with Noise}
     \end{subfigure}
     \caption{\FNN\ at high level of image corruption ($=0.6$).}
     \label{fig:fastNN_high_corruption_06}
\end{figure*}

\begin{figure*}[ht]
     \centering
     \begin{subfigure}{\textwidth}
         \centering
	 	 \includegraphics[width=\textwidth]{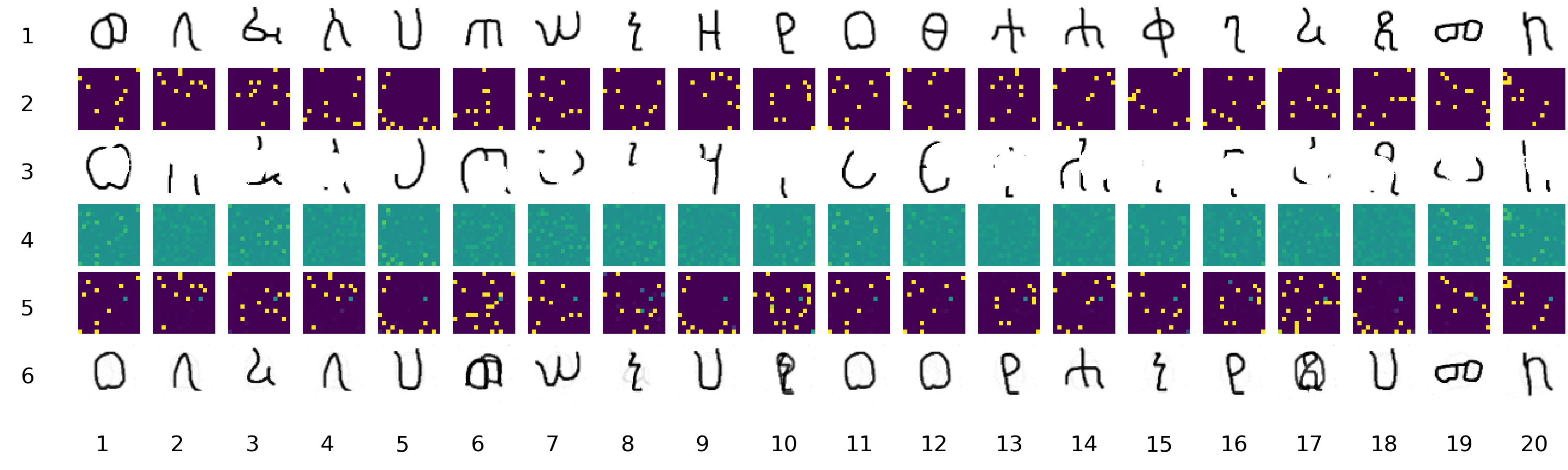}
		 \caption{\OneshotGen\ and Occlusion}
     \end{subfigure}
     \hfill
     \begin{subfigure}{\textwidth}
         \centering
	 	 \includegraphics[width=\textwidth]{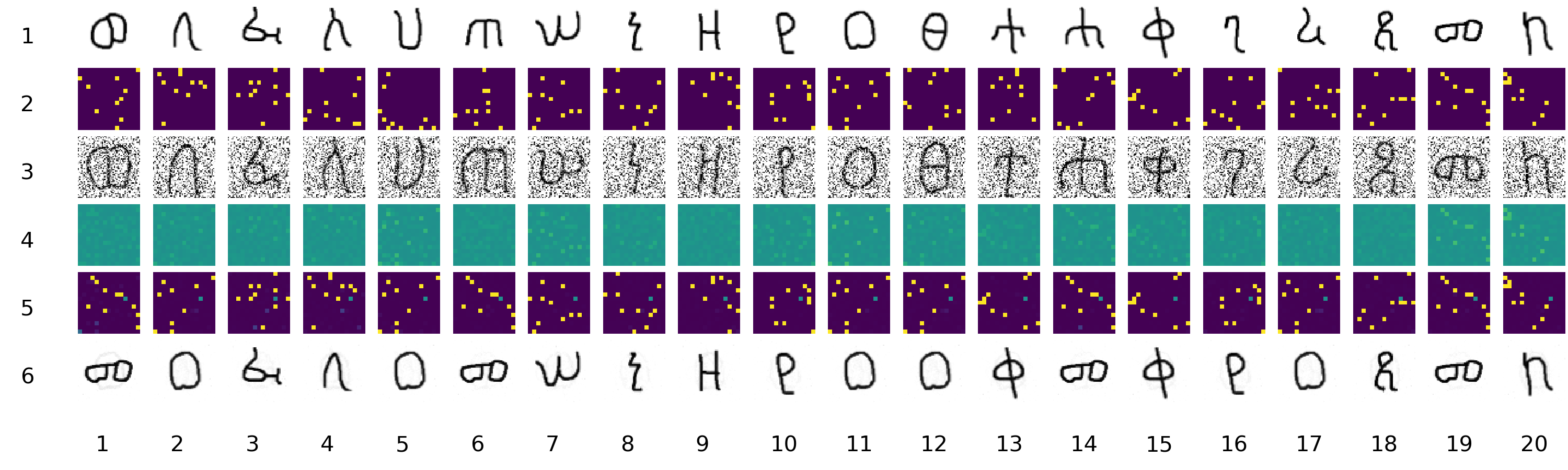}
  		 \caption{\OneshotGen\ with Noise}
     \end{subfigure}
     \begin{subfigure}{\textwidth}
         \centering
	 	 \includegraphics[width=\textwidth]{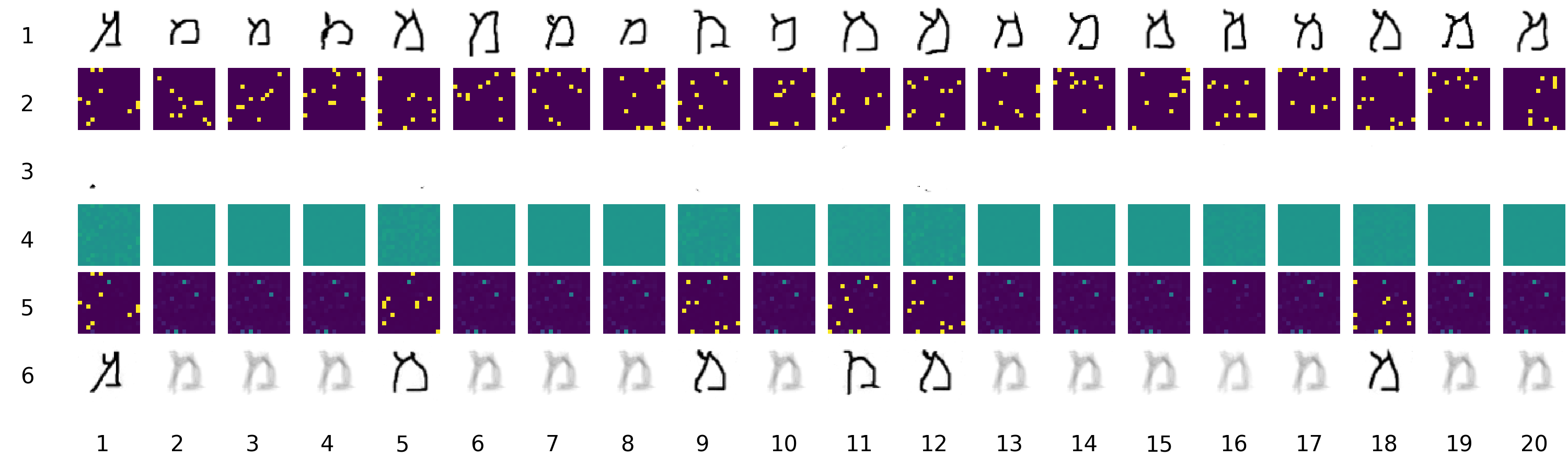}
		 \caption{\OneshotInst\ with Occlusion}
     \end{subfigure}
     \hfill
     \begin{subfigure}{\textwidth}
         \centering
	 	 \includegraphics[width=\textwidth]{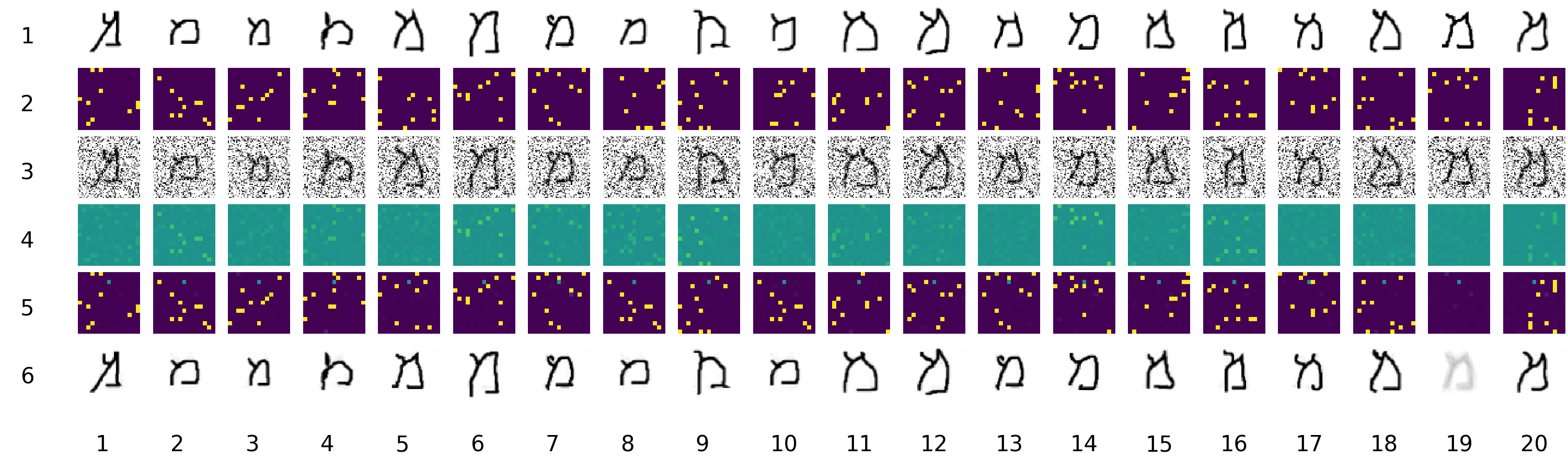}
  		 \caption{\OneshotInst\ with Noise}
     \end{subfigure}
     \caption{AHA at extreme level of image corruption ($=0.9$).}
     \label{fig:AHA_high_corruption_09}
\end{figure*}

\begin{figure*}[ht]
     \centering
     \begin{subfigure}{\textwidth}
         \centering
	 	 \includegraphics[width=\textwidth]{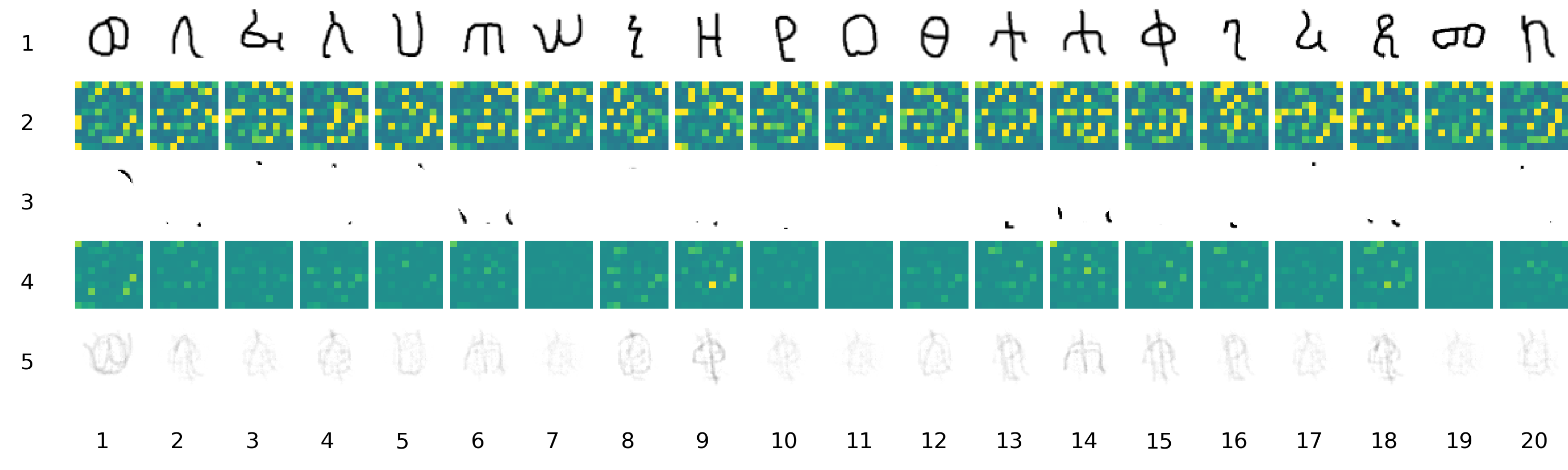}
		 \caption{\OneshotGen\ with Occlusion}
     \end{subfigure}
     \hfill
     \begin{subfigure}{\textwidth}
         \centering
	 	 \includegraphics[width=\textwidth]{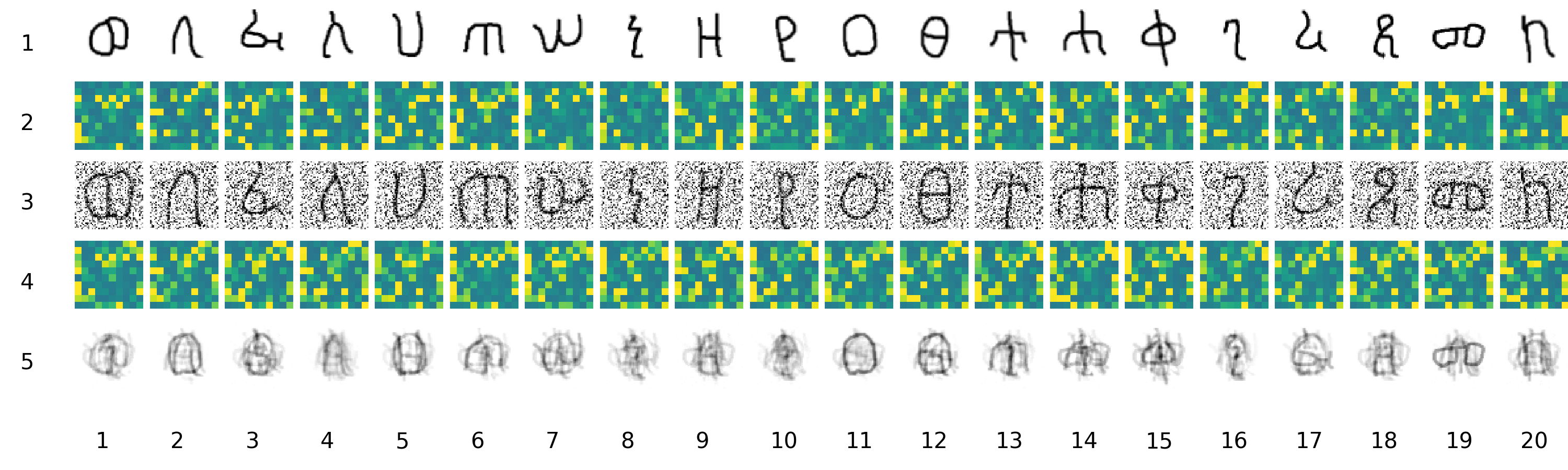}
  		 \caption{\OneshotGen\ with Noise}
     \end{subfigure}
     \begin{subfigure}{\textwidth}
         \centering
	 	 \includegraphics[width=\textwidth]{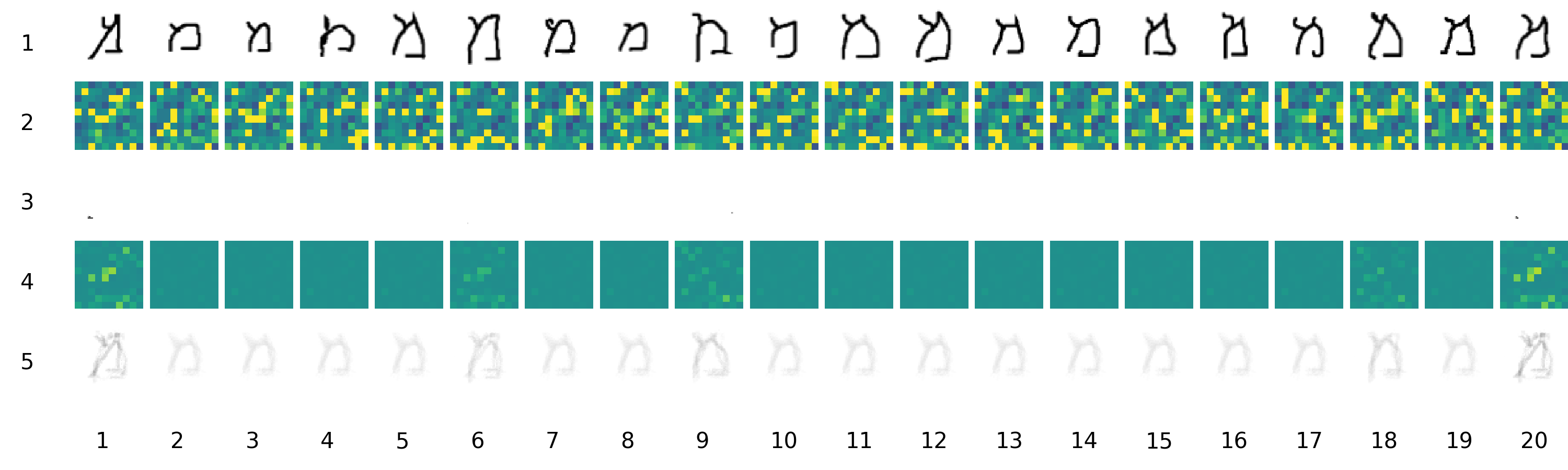}
		 \caption{\OneshotInst\ with Occlusion}
     \end{subfigure}
     \hfill
     \begin{subfigure}{\textwidth}
         \centering
	 	 \includegraphics[width=\textwidth]{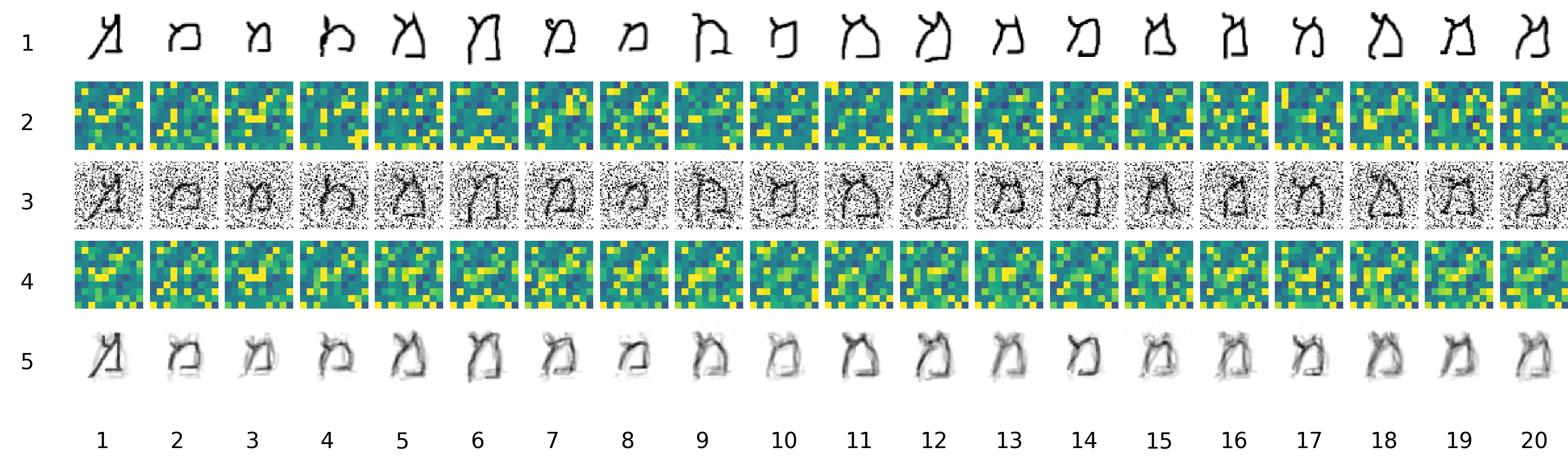}
  		 \caption{\OneshotInst\ with Noise}
     \end{subfigure}
     \caption{\FNN\ at extreme level of image corruption ($=0.9$).}
     \label{fig:fastNN_high_corruption_09}
\end{figure*}

High and extreme levels of image corruption are shown in Figures \ref{fig:AHA_high_corruption_06} to \ref{fig:fastNN_high_corruption_09}.

\end{document}